\documentclass[11pt]{article}

\usepackage[preprint]{acl}

\usepackage{times}
\usepackage{latexsym}
\usepackage{rotating} %
\usepackage[T1]{fontenc}

\usepackage[utf8]{inputenc}

\usepackage{microtype}

\usepackage{inconsolata}

\usepackage{graphicx}
\usepackage{amsmath}
\usepackage{amssymb}
\usepackage[ruled,vlined]{algorithm2e}
\usepackage{adjustbox}
\usepackage{booktabs}
\usepackage{caption}
\usepackage{enumitem}
\usepackage{float}
\usepackage[T1]{fontenc}
\usepackage{longtable}
\usepackage{linguex}
\usepackage{listings}
\usepackage{multirow}
\usepackage{natbib}
\usepackage{soul}
\usepackage{subcaption}
\usepackage{relsize}
\usepackage{tabularray}
\usepackage{tikz}
\usepackage{todonotes}
\usepackage{xcolor}
\usepackage{comment}

\lstdefinestyle{promptbox}{
  basicstyle=\ttfamily\small,
  frame=single,
  breaklines=true,
  columns=fullflexible,
  tabsize=2
}

\title{Finding Culture-Sensitive Neurons in Vision-Language Models}

\author{Xiutian Zhao$^{1,3} $ \quad Rochelle Choenni$^{2}$ \quad Rohit Saxena$^{1}$ \quad Ivan Titov$^{1,2}$ \\
  $^{1}$University of Edinburgh \quad $^{2}$University of Amsterdam \quad $^{3}$Johns Hopkins University
}

\begin{document}

\maketitle

\begin{abstract}
Despite their impressive performance, vision-language models (VLMs) still struggle on culturally situated inputs. 
To understand how VLMs process culturally grounded information, we study the presence of culture-sensitive neurons, i.e., neurons whose activations show preferential sensitivity to inputs associated with particular cultural contexts.
We examine whether such neurons are important for culturally diverse visual question answering and where they are located. Using the CVQA benchmark, we identify neurons of culture selectivity and perform diagnostic tests by deactivating the neurons flagged by various identification methods. Experiments on three VLMs across 25 cultural groups demonstrate the existence of neurons whose ablation disproportionately harms performance on questions about the corresponding cultures, while having limited effects on others. 
Moreover, we introduce a new margin-based
selector---Contrastive Activation Margin (ConAct)---and show that it outperforms probability- and entropy-based methods in identifying neurons associated with cultural selectivity. 
Finally, our layer-wise analyses reveal that such neurons are not uniformly distributed: they cluster in specific decoder layers in a model-dependent way.
\footnote{Related code is available at \url{https://github.com/xiutian/vlm-culture-neuron}.}
\end{abstract}

\section{Introduction}
\label{sec:intro}

Vision-language models (VLMs) underpin many multimodal applications, from visual question answering (VQA) to chart captioning and document parsing \cite{liu2023visual, li2023blip, Qwen2.5-VL, yue2025pangea}. Despite impressive performance, various works show that many VLMs struggle on culturally grounded visual content or culturally marked linguistic cues, and often exhibit systematic performance disparities across cultures \cite{cvqa, culturalvqa}. Understanding how and where such culture-related knowledge is represented within VLMs is important both for interpretability and fairness. %
Identifying subcomponents that are important for culture-related processing can not only improve our understanding of the underlying mechanisms, but may also guide future efforts to enhance these capabilities during post-training, e.g., through sparse fine-tuning \cite{ansell-etal-2022-composable, ben-zaken-etal-2022-bitfit} or activation steering \cite{turner2024steeringlanguagemodelsactivation, rimsky-etal-2024-steering}.

\begin{figure*}[t!]
    \centering
    \includegraphics[width=\textwidth]{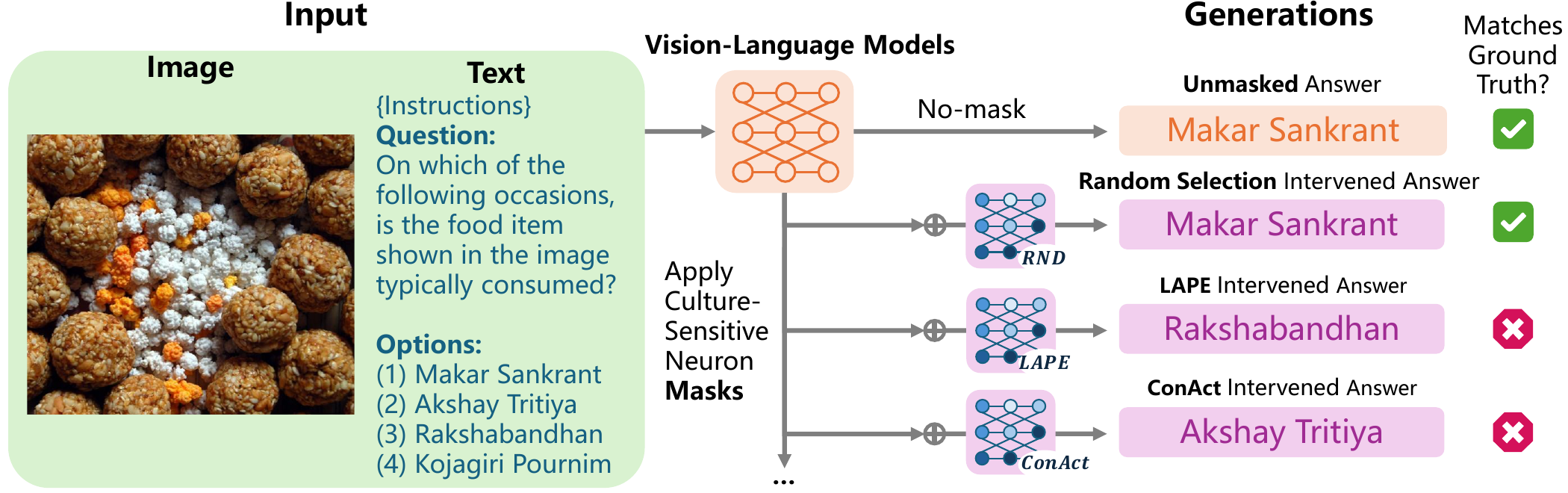}
    \caption{\textbf{An ablation example} of Qwen2.5-VL-7B on India-Marathi VQA subset. Given an image of Tilgul, an Indian sweet made from sesame seeds and jaggery, the full model selects the ground truth-matched option; \textbf{RND} mask does not affect the model's decision, while \textbf{LAPE} and \textbf{ConAct} masks redirect to different answers. Mentioned methods are explained in \S~\ref{subsec:identification}.
}
\label{fig:example}
\end{figure*}

Prior work in neural network interpretability has shown that individual neurons can exhibit relative specialization for certain concepts, modalities, or tasks \cite{bau2017networkdissectionquantifyinginterpretability, Bau_2020}.
In large language models (LLMs), researchers have found neurons that are preferentially active for particular languages \cite{tang-etal-2024-language}, knowledge domains \cite{yu-ananiadou-2024-neuron} and text-styles \cite{lai-etal-2024-style}. 
Analyses of VLMs, however, have primarily focused on modality-related aspects when identifying neuron functions (e.g., distinguishing neurons involved in visual vs. textual processing) \cite{huang2024minerminingunderlyingpattern, fang2024towards, xu2025deciphering}, leaving other forms of specialization underexplored. 
Specifically, it is unknown whether VLMs contain neurons that preferentially respond to inputs from specific cultural contexts, as opposed to comparable inputs from others. 
This question is especially relevant given that culture-related signals often arise from interactions between the visual and textual modalities. Addressing this gap can shed new light on how VLMs encode culturally grounded knowledge and where possible limitations or biases originate.

Thus, we study whether VLMs contain neurons whose activity is selectively modulated by culturally grounded inputs, without implying that these neurons are exclusively dedicated to culture. Instead, we aim to identify neurons that show relative culture-selectivity, i.e.,\ units whose activations exhibit stronger association with certain cultural contexts compared to others, and to evaluate to what extent such neurons are critical for culture-specific performance.
Concretely, we address the following questions: (1) Do VLMs contain such culture-sensitive neurons, i.e., neurons that preferentially activate on inputs tied to particular cultures? 
(2) Does ablating small, targeted subsets of these neurons selectively degrade a VLM's performance on questions tied to the corresponding culture, with minimal impact on other cultures? (3) How are these neurons distributed across layers, and is the pattern consistent across model architectures and cultures?

Following prior work on neuron detection \cite{tang-etal-2024-language, huo-etal-2024-mmneuron, huang2024minerminingunderlyingpattern, fang2024towards}, we adapt activation-based neuron analysis to a multimodal setting and evaluate on the CVQA benchmark \cite{cvqa}, operationalizing culture via the CVQA taxonomy of country--language pairs. We conduct experiments on three VLMs : Qwen2.5-VL-7B \cite{Qwen2.5-VL}, LLaVA-v1.6-Mistral-7B \cite{liu2023visual}, and Pangea-7B (\citealp{yue2025pangea}),  across 25 cultures. To minimize influence from differences in language proficiency or language-correlated effects, we constrain the experiments to a monolingual (English) setting. Moreover, to better isolate culture-sensitive neurons, we introduce \emph{Contrastive Activation Margin} (ConAct), a margin-based method that rewards large separation between a neuron's activation for its top-responding culture and its nearest competing culture, improving upon existing probability- and entropy-based selectors.

We provide empirical evidence for the existence of culture-sensitive neurons in VLMs. Ablating these neurons disproportionately reduces model performance on questions tied to the corresponding culture while leaving others largely unaffected, suggesting a causal role in culturally grounded information processing. Moreover, our layer-wise analysis reveals that these neurons are distributed across the decoder, with noticeable concentrations in mid-to-late layers. While we do observe some exceptions, this pattern remains largely consistent across the VLMs and cultures we examine. Overall, our results provide insight into how VLMs represent cultural knowledge and suggest new avenues for targeted evaluation and intervention to mitigate cultural biases or steer model behavior.

\section{Related work}\label{sec:background}
\begin{figure*}[t]
    \centering
    \includegraphics[width=\textwidth]{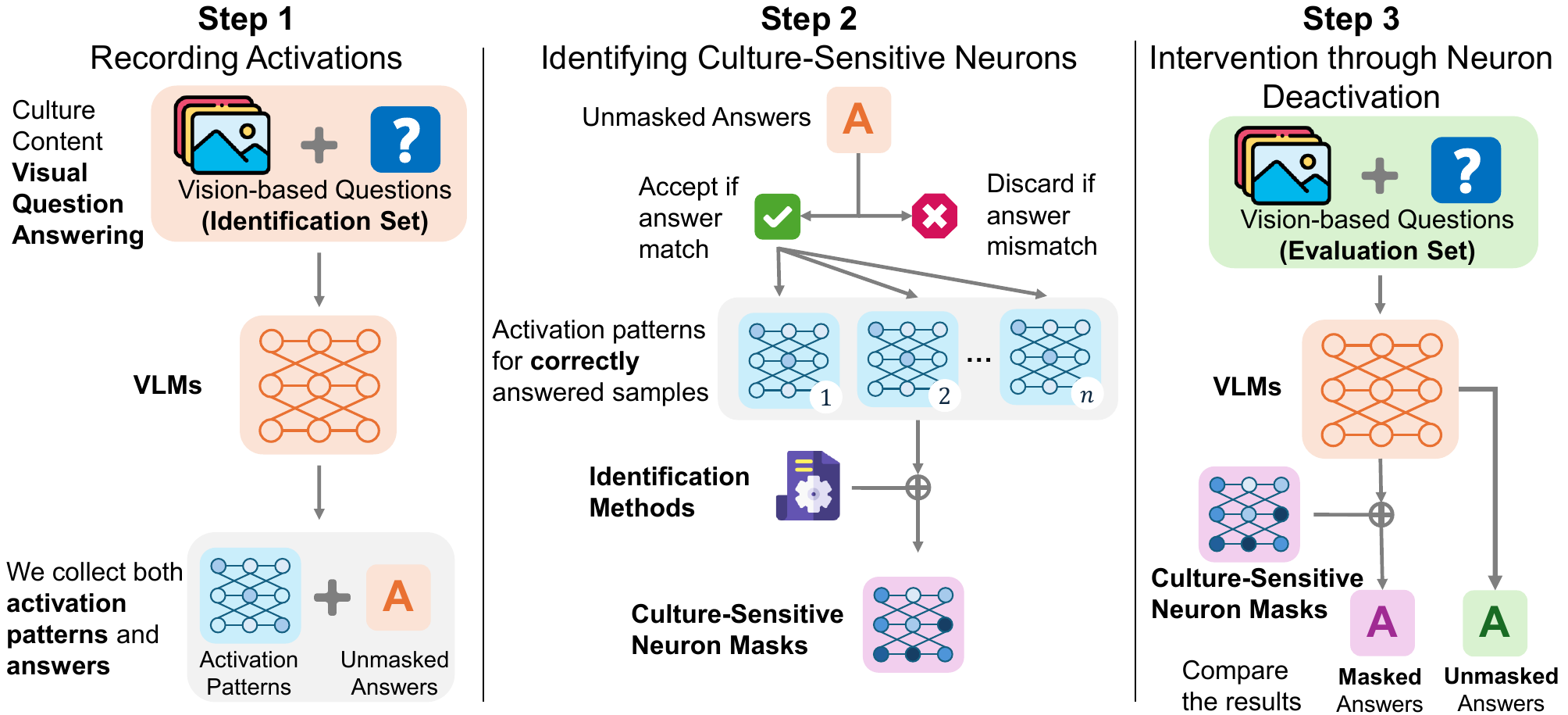}
    \caption{Pipeline for identifying and validating culture-sensitive neurons: 
    (1) record neuron activations on culture-specific VQAs, 
    (2) identify influential neurons using several methods, and 
    (3) evaluate their importance by ablating the top-$r\%$ neurons and measuring the effect on accuracy and answer divergence.}
    \label{fig:pipeline}
\end{figure*}

\paragraph{Studying neuron specialization.}
Identifying specialized neurons that respond strongly to particular features or concepts is a well-established practice in interpreting deep neural network models. Early work on convolutional neural network interpretability \cite{bau2017networkdissectionquantifyinginterpretability, Bau_2020} showed that individual hidden units can align with human-understandable concepts, such as objects, parts, colors, or even high-level concepts.
Analogous analyses have been applied to modern LLMs. For instance, \citet{yu-ananiadou-2024-neuron} showed potential neurons specialized at domain-knowledge; \citet{tang-etal-2024-language} introduced an entropy-based method to find language-specific neurons. Two relevant works demonstrated evidence of culture-related neurons in LLMs \cite{namazifard-poech-2025-isolating, yamamoto2025neuronlevelanalysisculturalunderstanding}.
However, in multimodal settings, existing efforts are concentrated on identifying modality- \cite{pan-etal-2024-finding, huang2024minerminingunderlyingpattern, fang2024towards, xu2025deciphering} or task-specific neurons \cite{neo2025towards}, whereas domain-knowledge associated neurons are limited \cite{huo-etal-2024-mmneuron, zhao2026discoveringcausallyvalidatingemotionsensitive}.

\paragraph{Cultural knowledge and bias in VLMs.}
Culture is a complex, multifaceted construct involving shared knowledge, practices, symbols and social norms of a group \cite{tylor1871primitive, hofstede1980culture}.
Culture-related multimodal benchmarks, such as CVQA \cite{cvqa}, \textsc{CulturalVQA} \cite{culturalvqa}, and \textsc{CulturalGround} \cite{nyandwi-etal-2025-grounding}, approximate culture via local knowledge and practices that are common in a region or within a language group \cite{10.1162/COLI.a.14}.
Such datasets test VLMs on culturally diverse content (e.g., traditional foods, clothing, landmarks), often revealing VLMs' substantial performance disparities across different cultures. 
Moreover, prior studies have found that VLMs tend to exhibit systematic biases both in the image perception and natural language reasoning \cite{madasu2025cultural, burda2025culturally, ananthram2025see, yadav-etal-2025-beyond}.

\section{Methodology}
\label{sec:method}

Following prior work on activation-based neuron analysis \cite{huo-etal-2024-mmneuron, huang2024minerminingunderlyingpattern, fang2024towards, tang-etal-2024-language}, we use a three-stage pipeline (Figure~\ref{fig:pipeline}): 
(1) record decoder-MLP neuron activations on culturally grounded VQA items that the unablated model answers correctly;
(2) score and select culture-sensitive neurons using the identification methods in \S\ref{subsec:identification}; %
and (3) causally test these neurons by inference-time deactivation and measure culture-specific performance changes.

\subsection{Step 1: Recording Activations}
\label{subsec:recording}
We instrument the decoder MLPs of each VLM and record neuron activations on VQAs that the unmasked model answers correctly.
The assumption is that neurons that are preferentially active when processing information tied to a particular culture will display distinctive activation patterns on the respective culture's inputs.

\paragraph{SwiGLU activations.}
We focus on the decoder MLP nonlinearity branch in the SwiGLU block \cite{shazeer2020gluvariantsimprovetransformer}.
Let $h_{l-1,t}\in\mathbb{R}^{d}$ be the layer-$l\!-\!1$ hidden state at token position $t$.
A standard SwiGLU MLP computes
\begin{align}
u_{l,t} &= h_{l-1,t} W_u + b_u, \qquad
v_{l,t} = h_{l-1,t} W_v + b_v, \label{eq:swiglu_uv}\\
g_{l,t} &= \mathrm{SiLU}(u_{l,t}), \qquad
z_{l,t} = (g_{l,t}\odot v_{l,t}) W_o + b_o, \label{eq:swiglu_out}
\end{align}
where $\mathrm{SiLU}(x)=x\,\sigma(x)$ and $\sigma(\cdot)$ is the sigmoid.
For each neuron (dimension) $n$ in layer $l$ and token $t$, we denote the recorded scalar activation by
\begin{equation}
a_{l,n,t} \;=\; (g_{l,t})_n .
\end{equation}
Because $\sigma(x)>0$ for all $x$, $\mathrm{SiLU}(x)$ has the same sign as $x$, so $\mathbb{I}(a_{l,n,t}>0)$ is equivalent to $\mathbb{I}((u_{l,t})_n>0)$.

\paragraph{Valid-token masking.}
Let $m_{i,t}\in\{0,1\}$ be a valid-token mask for example $i$ at token position $t$, consistent with each model's internal attention mask.
This excludes padding and special markers (e.g., image delimiters), while retaining both text tokens and visual tokens consumed by the decoder.
We instrument only the decoder (not upstream vision encoders).

\paragraph{Activation statistics.}
Let $\mathcal{C}$ be the set of cultures and let $\mathcal{I}_c$ be the set of correctly answered identification examples tagged with culture $c$.
We accumulate per-neuron, per-culture sufficient statistics:
\begin{align}
K_{l,n}^{(c)} &= \sum_{i\in \mathcal{I}_c}\sum_{t} m_{i,t}\,\mathbb{I}\!\big(a_{l,n,t}^{(i)} > 0\big), \label{eq:K_count}\\
T_{c} &= \sum_{i\in \mathcal{I}_c}\sum_{t} m_{i,t}. \label{eq:T_tokens}
\end{align}
Here $K$ counts how often a neuron fires positively, 
and $T_c$ is the total number of valid tokens observed for culture $c$.
Restricting to correctly answered examples reduces noise from activations associated with failures.

\subsection{Step 2: Identification of Culture-Sensitive Neurons}
\label{subsec:identification}
Using the aggregated statistics from Step 1, we define the token-level activation probability:
\begin{equation}
P_{l,n}^{(c)} \;=\; \frac{K_{l,n}^{(c)}}{T_c}. 
\label{eq:mean-mag}
\end{equation}

\subsubsection{Baseline Identification Methods}
As our identification methods for neuron scoring, we consider the following existing baseline methods. For each culture $c$, these methods return a ranking of neuron indices from most to least selective. Deciding how many of those neurons to select as culture-sensitive is a hyperparameter setting. To allow for a fair comparison across methods, we select the $r\%$ highest scoring neurons out of all MLP-neurons as culture-sensitive, where we set $r{=}1$ throughout.

\paragraph{Random Selection (RND)}
To evaluate if cultural subsets are inherently sensitive to arbitrary masking, we use a global random baseline that samples a fixed total number of neurons (i.e., r\%) uniformly from all layers, independent of culture.
This produces a single mask shared across cultures and is compute-efficient, without enforcing any layer-wise quota.

\paragraph{Activation Probability (LAP)} \cite{gurnee2024universal, voita-etal-2024-neurons}.
LAP selects neurons that frequently fire for a given target (here adapted to culture), which emphasizes firing frequency alone. Directly using the activation probability:
\begin{equation}
s^{\mathrm{LAP}}_{l,n}(c) \;=\; P_{l,n}^{(c)}.
\end{equation}
For each $c$, we rank neurons by $s^{\mathrm{LAP}}_{l,n}(c)$ and select the top $r\%$.

\paragraph{Activation Probability Entropy (LAPE)} \cite{tang-etal-2024-language, huo-etal-2024-mmneuron, namazifard-poech-2025-isolating}.
LAPE measures how selectively a neuron fires across cultures by computing the entropy of its culture profile.
For each neuron, define the (approximately) normalized culture profile:
\begin{equation}
\tilde{P}^{(c)}_{l,n}
=
\frac{P^{(c)}_{l,n}}{\sum_{c'\in\mathcal{C}} P^{(c')}_{l,n}+\epsilon},
\quad \epsilon>0,
\end{equation}
and Shannon entropy:
\begin{equation}
s^{\mathrm{LAPE}}_{l,n}
=
-\sum_{c\in\mathcal{C}}\tilde{P}^{(c)}_{l,n}\log \tilde{P}^{(c)}_{l,n}.
\end{equation}
Because $\epsilon$ is used only for numerical stability, $\tilde{P}_{l,n}$ is nearly normalized for active neurons (where $\sum_{c'} P^{(c')}_{l,n}\gg \epsilon$).
Lower entropy indicates stronger culture selectivity. 

We implement the following procedure:
(1) \textbf{Activity filter.} We keep neurons whose maximal firing probability exceeds a threshold: $\max_{c} P^{(c)}_{l,n} > p_{\mathrm{th}}$, where $p_{\mathrm{th}}$ is set to the $\alpha$-percentile of all values in $\{P^{(c)}_{l,n}\}$ (we use $\alpha=95$). Neurons failing this criterion are treated as inactive and excluded from candidate selection.
(2) \textbf{Low-entropy candidate pool.} From the remaining neurons, we select a candidate pool consisting of the lowest-$\rho$ fraction by $s^{\mathrm{LAPE}}_{l,n}$. In our implementation, we set $\rho=\min(1,5r)$, i.e., the candidate pool size is at most five times the final selection rate.
(3) \textbf{Per-culture selection.} For each culture $c$, we select the top-$r\%$ neurons within the candidate pool by their firing probabilities $P^{(c)}_{l,n}$ to form the culture-specific mask. (A neuron may be selected for multiple cultures because masks are constructed independently per culture.)

\paragraph{MAD (Mean Activation Difference)} \cite{bau2018identifying, 10.1609/aaai.v33i01.33016309}.
We use the same post-nonlinearity activations $a^{(i)}_{l,n,t}$ and valid-token mask $m_{i,t}$ as in Step~1.
Define the across-culture mean firing probability:
\begin{equation}
\bar{P}_{l,n} = \frac{1}{|C|}\sum_{c\in C} P^{(c)}_{l,n}.
\end{equation}
The MAD score for culture $c$ is the absolute deviation from this mean:
\begin{equation}
s^{\mathrm{MAD}}_{l,n}(c) = \left|P^{(c)}_{l,n} - \bar{P}_{l,n}\right|.
\end{equation}
For each culture $c$, we rank neurons by $s^{\mathrm{MAD}}_{l,n}(c)$ and select the top $r\%$.

\subsubsection{Contrastive Activation Margin (ConAct)}
\label{subsec:ccs}
In a preliminary analysis, we compute each neuron's standard deviation of $P^{(c)}_{l,n}$ across cultures and observe that in Qwen2.5-VL-7B and Pangea-7B, a substantial fraction of neurons (12.27\% and 9.57\%) satisfy $\mathrm{std}_c(P^{(c)}_{l,n}) > \mathrm{mean}_c(P^{(c)}_{l,n})$ (Appendix~\ref{appendix:full_unmasked} Table~\ref{table:mad_stat}), exhibiting high activation variance across cultures.
This suggests that a large mean-based difference may not necessarily indicate cultural specialization but may arise from high intrinsic variability.

To mitigate this, we introduce \emph{Contrastive Activation Margin} (ConAct), a margin-based selector that measures the gap between the most-activated culture and its nearest competitor.
By focusing on this contrast rather than deviation from the mean, ConAct is less sensitive to global variance and is expected to be more effective in high-variance models.
We thus hypothesize that deactivating ConAct-identified neurons will lead to a larger culture-specific performance drop in such models, while in low-variance models, ConAct and MAD will likely identify similar neurons.
For each neuron $(l,n)$, define the top culture and runner-up by activation probability:
\begin{align}
c^{(1)}_{l,n} &= \arg\max_{c\in\mathcal{C}} P_{l,n}^{(c)}, \qquad
P^{(1)}_{l,n} = \max_{c\in\mathcal{C}} P_{l,n}^{(c)},\\
P^{(2)}_{l,n} &= \max_{c\in\mathcal{C}\setminus\{c^{(1)}_{l,n}\}} P_{l,n}^{(c)} .
\end{align}
The ConAct score assigns each neuron exclusively to its top culture:
\begin{equation}
s^{\mathrm{ConAct}}_{l,n}(c) \;=\;
\begin{cases}
P^{(1)}_{l,n} - P^{(2)}_{l,n}, & \text{if } c=c^{(1)}_{l,n},\\
-\infty, & \text{otherwise.}
\end{cases}
\label{eq:conact_score}
\end{equation}

If we use an aggregated culture group (e.g., pooling multiple country--language pairs), we treat it as a single culture in $C$ by pooling its member pairs at the data level (i.e., $I_c$ is the union of examples in that group). All sufficient statistics are then computed on the pooled set. ConAct is applied unchanged on the resulting $\{P^{(c)}_{l,n}\}_{c\in C}$.

\subsection{Step 3: Intervention through Neuron Deactivation}
\label{subsec:intervention}
We causally test whether neurons selected in Step 2 are important for culture-sensitive behavior by deactivating them at inference time and measuring the impact on the evaluation split.

Let $\mathcal{M}^{(m,c_{\mathrm{src}})}_l \subseteq \{1,\dots,D_l\}$ be the set of neuron indices selected by method $m$ for source culture $c_{\mathrm{src}}$ at decoder layer $l$ (where $D_l$ is the SwiGLU hidden width).
We form a binary keep-mask $r^{(m,c_{\mathrm{src}})}_l\in\{0,1\}^{D_l}$:
\begin{equation}
r^{(m,c_{\mathrm{src}})}_{l,n} \;=\;
\begin{cases}
0, & n\in \mathcal{M}^{(m,c_{\mathrm{src}})}_l \quad\text{(deactivate)}\\
1, & \text{otherwise.}
\end{cases}
\label{eq:keepmask}
\end{equation}
During inference, we apply this mask to the SwiGLU nonlinearity output (broadcast over tokens):
\begin{equation}
\tilde g_{l,t} \;=\; g_{l,t}\odot r^{(m,c_{\mathrm{src}})}_l,
\end{equation}
and replace Eq.~\eqref{eq:swiglu_out} with
\begin{equation}
z_{l,t} \;=\; (\tilde g_{l,t}\odot v_{l,t})W_o + b_o,
\end{equation}
leaving all other components unchanged.
We then compare masked vs.\ unmasked generations on each evaluation culture to quantify culture-specific effects.

\section{Experimental Setup}
\label{sec:exp-setup}

\subsection{Dataset and Culture Grouping}
We employ the CVQA dataset \cite{cvqa} as our testbed and operationalize ``culture'' through CVQA's country--language pair (e.g., ``Ireland--Irish'') taxonomy. 
Each item is a VQA question paired with an image and tagged by a country--language pair. 
We treat most country--language pairs as standalone cultures. Additionally, to study grouping effects, we form three aggregated culture groups by pooling pairs that share a country tag (India-all; Indonesia-all) or a language tag (all-Spanish). A subset of the culture groups and their question counts is shown in Table~\ref{table:grouping} (full list in Appendix~\ref{subsec:grouping}).

To minimize confounding from language proficiency, we use the dataset's prepared English translations for both questions and answer options. Moreover, this mitigates the concern of identifying language rather than culture-sensitive neurons.
We split the dataset approximately 50/50 into identification and evaluation subsets, where the identification split is used exclusively for activation logging (Step~1) and the evaluation split for masked generation and evaluation (Step~3).

\begin{table}[t!]
\centering
\resizebox{0.98\columnwidth}{!}{%
\begin{tabular}{@{}lccc@{}}
\toprule
CVQA Pairs            & Cultures     & \# Qs (I) & \# Qs (E) \\ \midrule
Brazil--Portuguese     & BRA                  & 142       & 142       \\
Bulgaria--Bulgarian    & BGR                  & 185       & 186       \\
China--Chinese         & CHN                  & 155       & 156       \\
Egypt--Egyptian Arabic & EGY                  & 101       & 102       \\
Ethiopia--Amharic      & ETA                  & 117       & 117       \\
Ethiopia--Oromo        & ETO                  & 107       & 107       \\
France--Breton         & FRA                  & 202       & 203       \\ \midrule
India--Bengali         & \multirow{6}{*}{IND} & 143       & 143       \\
India--Hindi           &                      & 100       & 101       \\
India--Marathi         &                      & 101       & 101       \\
India--Tamil           &                      & 107       & 107       \\
India--Telugu          &                      & 100       & 100       \\
India--Urdu            &                      & 110       & 110       \\ \midrule
Indonesia--Indonesian  & \multirow{4}{*}{IDN} & 206       & 206       \\
Indonesia--Javanese    &                      & 148       & 149       \\
Indonesia--Minangkabau &                      & 125       & 126       \\
Indonesia--Sundanese   &                      & 100       & 100       \\ \midrule
Ireland--Irish         & IRL                  & 163       & 163       \\
...                   & ...                  & ...       & ...       \\  \midrule
Total                 &                      & 5178      & 5196      \\ \bottomrule
\end{tabular}%
}
\caption{\textbf{Culture subset and VQA statistics}. CVQA country--language pairs with [``India'', ``Indonesia''] country tag are assigned to one of the grouped cultures. \textbf{\# Qs (I)} denotes the number of questions used for activation recording and neuron identification, while \textbf{\# Qs (E)} denotes the number of questions used for masked generation. Full table in Appendix~\ref{subsec:grouping}.}
\label{table:grouping}
\end{table}

\subsection{Models}
We evaluate three widely used VLMs: (1) LLaVA-v1.6-Mistral-7B \cite{liu2023visual, jiang2023mistral7b}, (2) Pangea-7B \cite{yue2025pangea}, and (3) Qwen2.5-VL-7B \cite{Qwen2.5-VL} (versions and sources can be found in Appendix~\ref{appendix:reproduce} Table~\ref{table:models}).

The selected models differ in backbone, supervision, and cultural/linguistic coverage, allowing us to test whether culture-sensitive neurons emerge consistently across architectures and training paradigms. 
Moreover, we selected Pangea-7B because it was developed to support broad multilingual multimodal coverage, making it a natural candidate for evaluating culture-linked behavior across diverse regions.

\subsection{Prompting and Decoding}
We use a fixed multiple-choice instruction template (Appendix~\ref{appendix:prompt}) for all models, requiring the output to be the complete option content rather than the label.
Maximum generation length is set to 20, which is sufficient to return a full answer-option span.
Decoding is deterministic (temperature $0$; no sampling).
Generations violating the format are normalized by the extraction heuristic (Appendix~\ref{appendix:normalize}).

\subsection{Measuring Cultural Sensitivity}
\label{subsec:measuring_cultural_sensitivity}
Using each neuron selector outlined in \S~\ref{sec:method}, we obtain a set of culture-sensitive neurons for each source culture $c_{\mathrm{src}} \in \mathcal{C}$.
To evaluate to what extent these neurons are indeed culture-sensitive, 
we study two conditions:
(1) \textbf{Self-deactivation:} $c_{\mathrm{src}} = c_{\mathrm{eval}}$, where the same culture from which the neurons were identified was used for evaluation.
(2) \textbf{Cross-deactivation:} $c_{\mathrm{src}} \neq c_{\mathrm{eval}}$, where the neurons were identified from a culture that differs from the one under evaluation.
This design allows us to test whether the selected neurons are primarily associated with a particular culture rather than affecting the model's overall capacity.

\begin{table*}[t!]
\centering
\resizebox{0.82\textwidth}{!}{%
\begin{tabular}{@{}lll|crrrr@{}}
\toprule
VLM & \multicolumn{1}{l}{Metric} 
     &          Eval. Setting     & RND & LAP & LAPE & MAD & ConAct \\ 
\midrule
\multirow{6}{*}{Qwen2.5-VL-7B} 
  & \multirow{3}{*}{Acc.\ $\Delta$}
      & Self-Deactivation               & $-$0.19 & $+$0.96 & $+$0.56 & $-$4.64 & \textbf{$-$5.52} \\
  &   & Cross-Deactivation Avg.         &  --   & $+$1.07 & \textbf{$+$0.61} & $-$1.31 & $-$0.64 \\
  &   & Self--Cross Gap                  &  --   & $-$0.08 & $-$0.05 & $-$3.33 & \textbf{$-$4.88}  \\
\cmidrule(l){2-8}
  &  \multirow{3}{*}{Flip Rate}
      & Self-Deactivation               & 4.66  & \textbf{17.05} & 4.64 & 12.03 & 12.61 \\
  &   & Cross-Deactivation Avg.         &  --   & 17.21 & \textbf{4.12} & 5.96 & 4.25 \\
  &   & Self--Cross Gap                  &  --   & $-$0.16 & $+$0.52 & $+$6.07 & \textbf{$+$8.36} \\
\midrule
\multirow{6}{*}{Pangea-7B} 
  & \multirow{3}{*}{Acc.\ $\Delta$}
      & Self-Deactivation               & 1.02  & $+$1.00 & $-$0.74 & $-$4.20 & \textbf{$-$4.33} \\
  &   & Cross-Deactivation Avg.         &  --   & $+$0.89 & \textbf{$-$0.37} & $-$1.34 & $-$0.72 \\
  &   & Self--Cross Gap                  &  --   & $+$0.11 & $-$0.38 & $-$2.86 & \textbf{$-$3.61} \\
\cmidrule(l){2-8}
  & \multirow{3}{*}{Flip Rate}
      & Self-Deactivation               & 6.45  & \textbf{24.10} & 6.52 & 13.55 & 12.99 \\
  &   & Cross-Deactivation Avg.         &  --   & 23.82 & \textbf{6.18} & 8.80 & 7.34 \\
  &   & Self--Cross Gap                  &  --   & $+$0.28 & $+$0.34 & $+$4.75 & \textbf{$+$5.65} \\
\midrule
\multirow{6}{*}{\begin{tabular}[c]{@{}l@{}}LLaVA-v1.6\\ -Mistral-7B\end{tabular}} 
  & \multirow{3}{*}{Acc.\ $\Delta$}
      & Self-Deactivation               & $-$0.50 & $-$2.50 & \textbf{$-$4.43} & $-$1.46 & $-$1.39 \\
  &   & Cross-Deactivation Avg.         &  --   & $-$2.74 & $-$4.44 & \textbf{$-$0.53} & $-$0.63 \\
  &   & Self--Cross Gap                  &  --   & $+$0.24 & $+$0.01 & \textbf{$-$0.93} & $-$0.76 \\
\cmidrule(l){2-8}
  & \multirow{3}{*}{Flip Rate}
      & Self-Deactivation               & 7.01  & \textbf{17.82} & 11.44 & 7.74 & 9.58 \\
  &   & Cross-Deactivation Avg.         &  --   & 17.74 & 11.28 & \textbf{6.56} & 7.63 \\
  &   & Self--Cross Gap                  &  --   & $+$0.08 & $+$0.15 & $+$1.18 & \textbf{$+$1.95} \\
\bottomrule
\end{tabular}%
}
\caption{\textbf{Ablation results on CVQA using culture-sensitive neurons selected by five identification methods.}
We report signed changes of accuracy change and flip rate relative to the unablated full model. Self-deactivation is measured on the culture used for identification (diagonal), and cross-deactivation on other cultures (off-diagonal).
RND (Random Selection) is not a culture-specific masking and hence does not distinguish ``self-'' or ``cross-'' results. We use the self--cross gap to summarize cultural specificity; larger negative gap indicates stronger culture-specific impact with less spillover. We bold the method with the smallest cross effect, largest self-effect and self--cross gap.}
\label{table:main_results}
\end{table*}

\paragraph{Metrics.}
We assess each condition using two complementary metrics: (1) \textbf{Accuracy change} ($\Delta$) measures the change in task performance between the full model and the masked model on a particular subset. Let $Acc_{\text{full}}$ and $Acc_{\text{masked}}$ denote accuracies reported in percentage (i.e., in $[0,100]$). We define the accuracy change as the percentage-point difference. (2) \textbf{Flip rate} is the proportion of items whose predicted answers differ from the full model, revealing decision shifts regardless of accuracy.
Concretely, let $\hat{a}_i^{\text{mask}}$ and $\hat{a}_i^{\text{full}}$ denote the model's predicted answers with and without ablation masking (after normalization; Appendix~\ref{appendix:normalize}) for item $i$, respectively.
\begin{align}
\nonumber
\Delta\mathrm{Acc} &= \mathrm{Acc}_{\text{masked}} - \mathrm{Acc}_{\text{full}}, \\
\nonumber
\mathrm{FlipRate} &= 100 \cdot \frac{1}{N}\sum_{i=1}^{N}\mathbb{I}\!\left[\hat{a}^{\text{full}}_i \neq \hat{a}^{\text{mask}}_i\right].
\end{align}

\paragraph{Interpretation.}
Ablating culture-sensitive neurons should harm performance when the evaluation culture matches the source culture (large negative $\Delta$, high flip rate), but have minimal effect otherwise ($\Delta$ and flip rate close to 0).
Hence, we focus on the gap between the self-deactivation effect and the average cross-deactivation effect as the main indicator of cultural sensitivity. Methods that yield larger self--cross gaps better isolate neurons that are critical, yet relatively specific to a given culture. 
Larger gaps (more negative) indicate stronger culture-specific impact with less spillover.

\section{Results}
\label{sec:results}
\begin{figure}[ht!]
    \centering
     \includegraphics[width=\columnwidth]{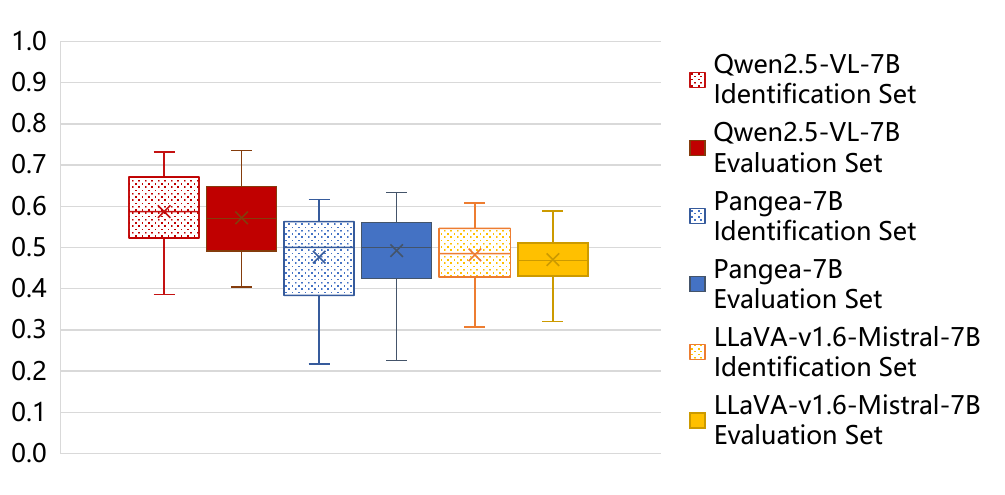}
    \caption{\textbf{Unablated full models per-culture accuracy on CVQA.} Distribution of per-culture accuracies for the three models on the identification split (marked in dots)  and the evaluation split (marked in solid color). The full table of per-culture results appears in Appendix~\ref{appendix:full_unmasked}.}
    \label{fig:train_box}
\end{figure}

\begin{figure*}[ht!]
    \centering
    \begin{subfigure}[b]{0.24\textwidth}
        \centering
        \includegraphics[width=\linewidth]{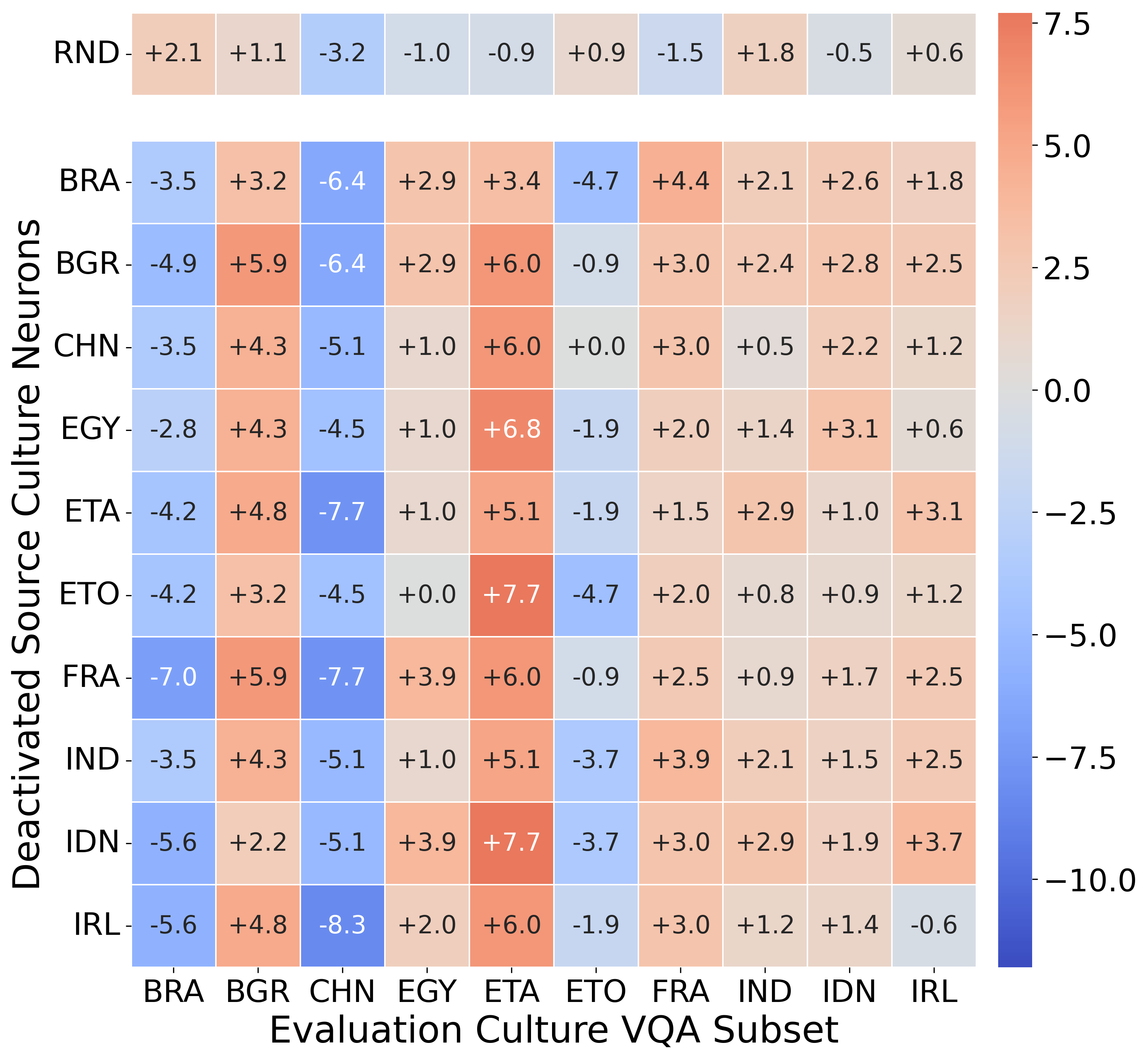}
    \end{subfigure}\hfill
    \begin{subfigure}[b]{0.24\textwidth}
        \centering
        \includegraphics[width=\linewidth]{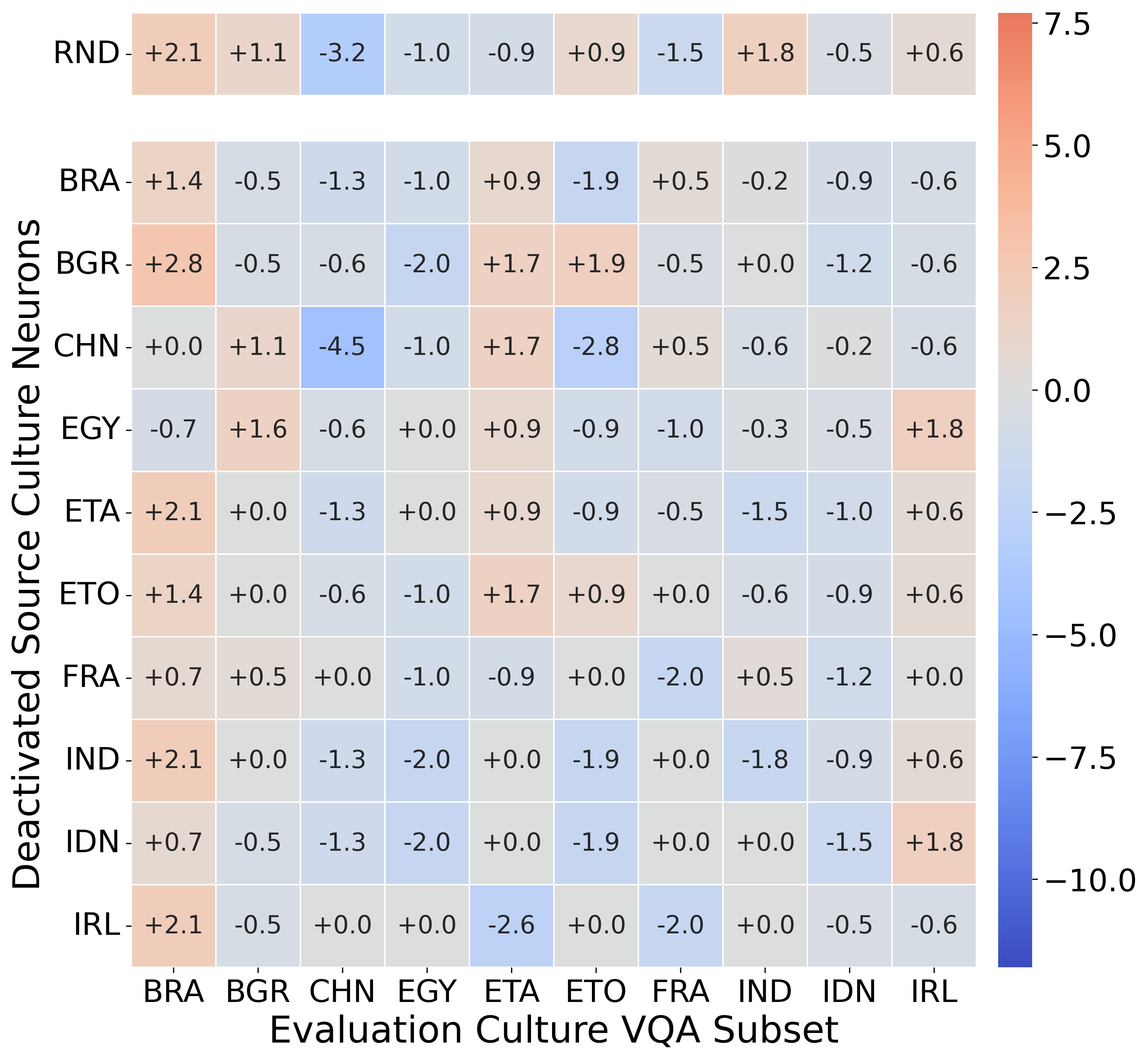}
    \end{subfigure}\hfill
    \begin{subfigure}[b]{0.24\textwidth}
        \centering
        \includegraphics[width=\linewidth]{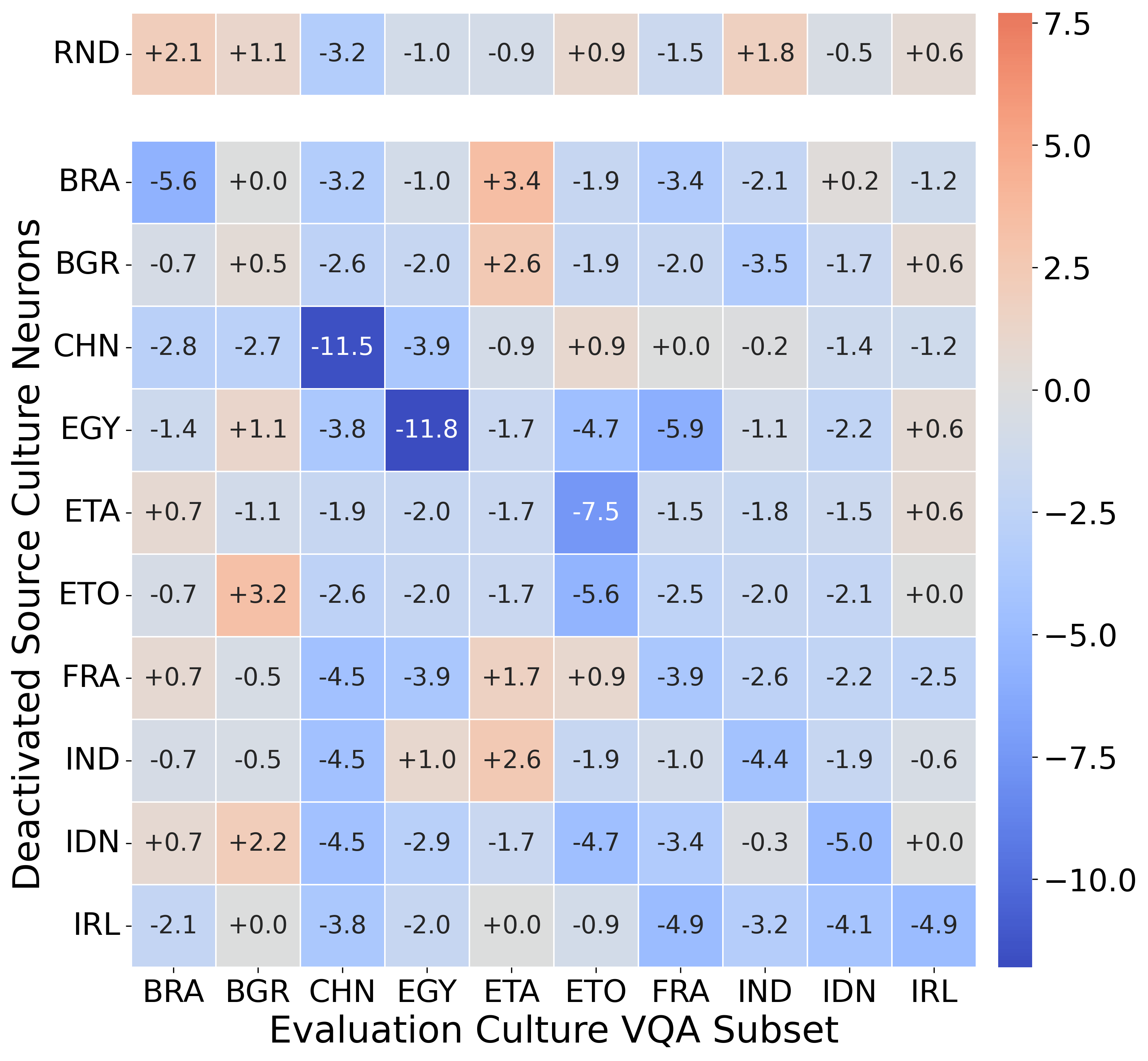}
    \end{subfigure}\hfill
    \begin{subfigure}[b]{0.24\textwidth}
        \centering
        \includegraphics[width=\linewidth]{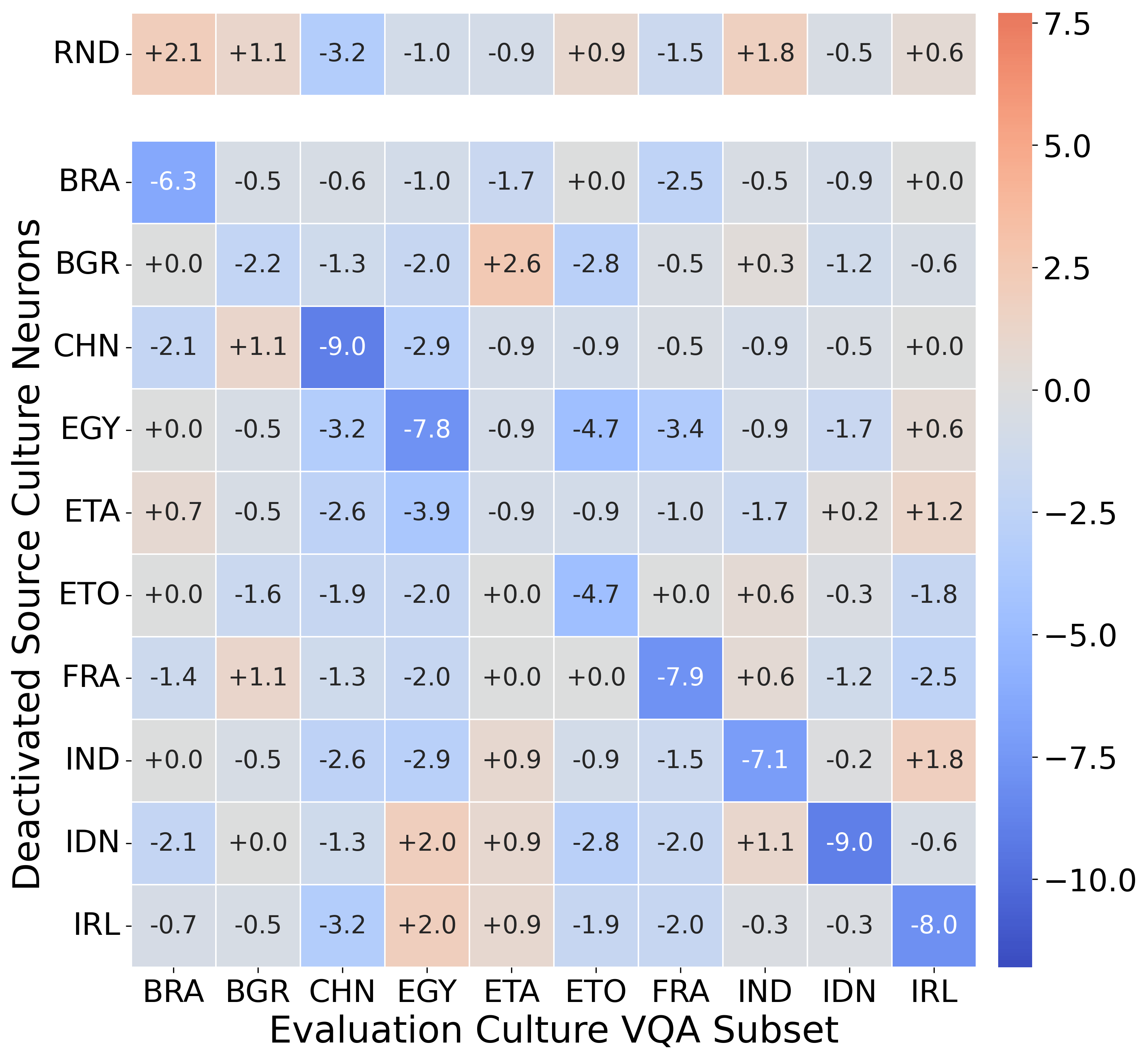}
    \end{subfigure}

    \vspace{1em} %

    \begin{subfigure}[b]{0.24\textwidth}
        \centering
        \includegraphics[width=\textwidth]{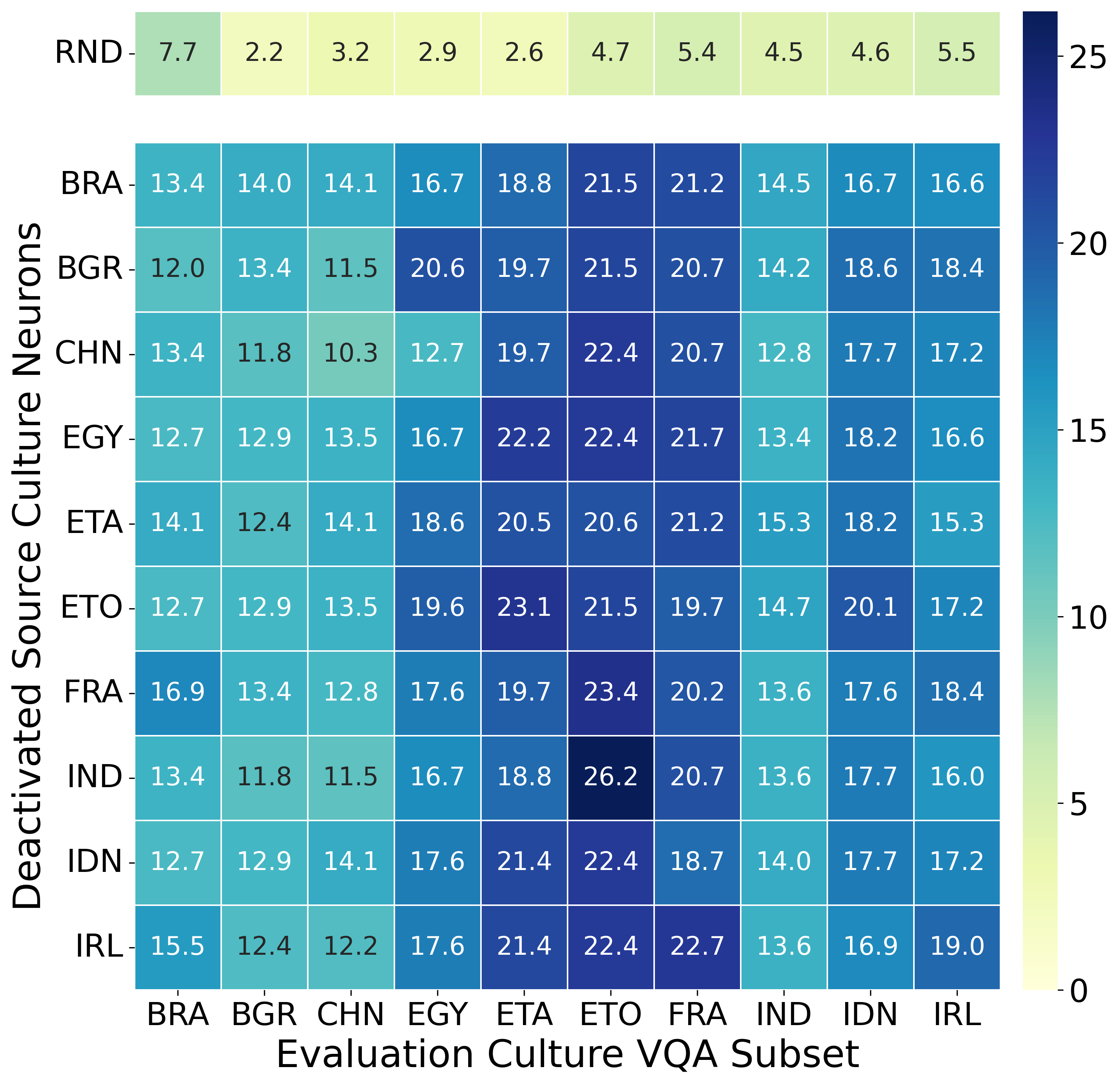}
        \caption{\textbf{LAP}}
    \end{subfigure}
    \hfill  
    \begin{subfigure}[b]{0.24\textwidth}
        \centering
        \includegraphics[width=\textwidth]{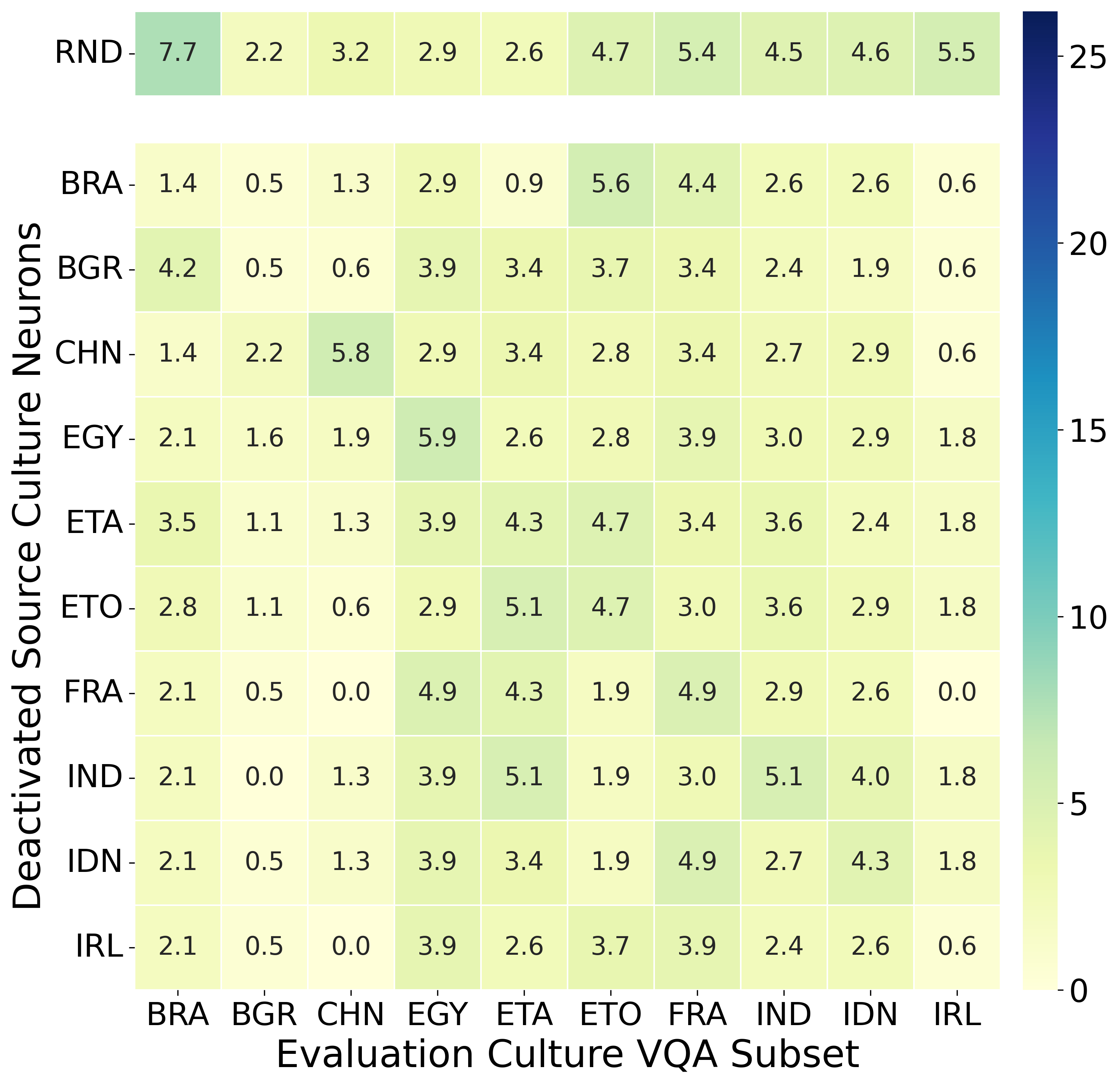}
        \caption{\textbf{LAPE}}
    \end{subfigure}
    \hfill
    \begin{subfigure}[b]{0.24\textwidth}
        \centering
        \includegraphics[width=\textwidth]{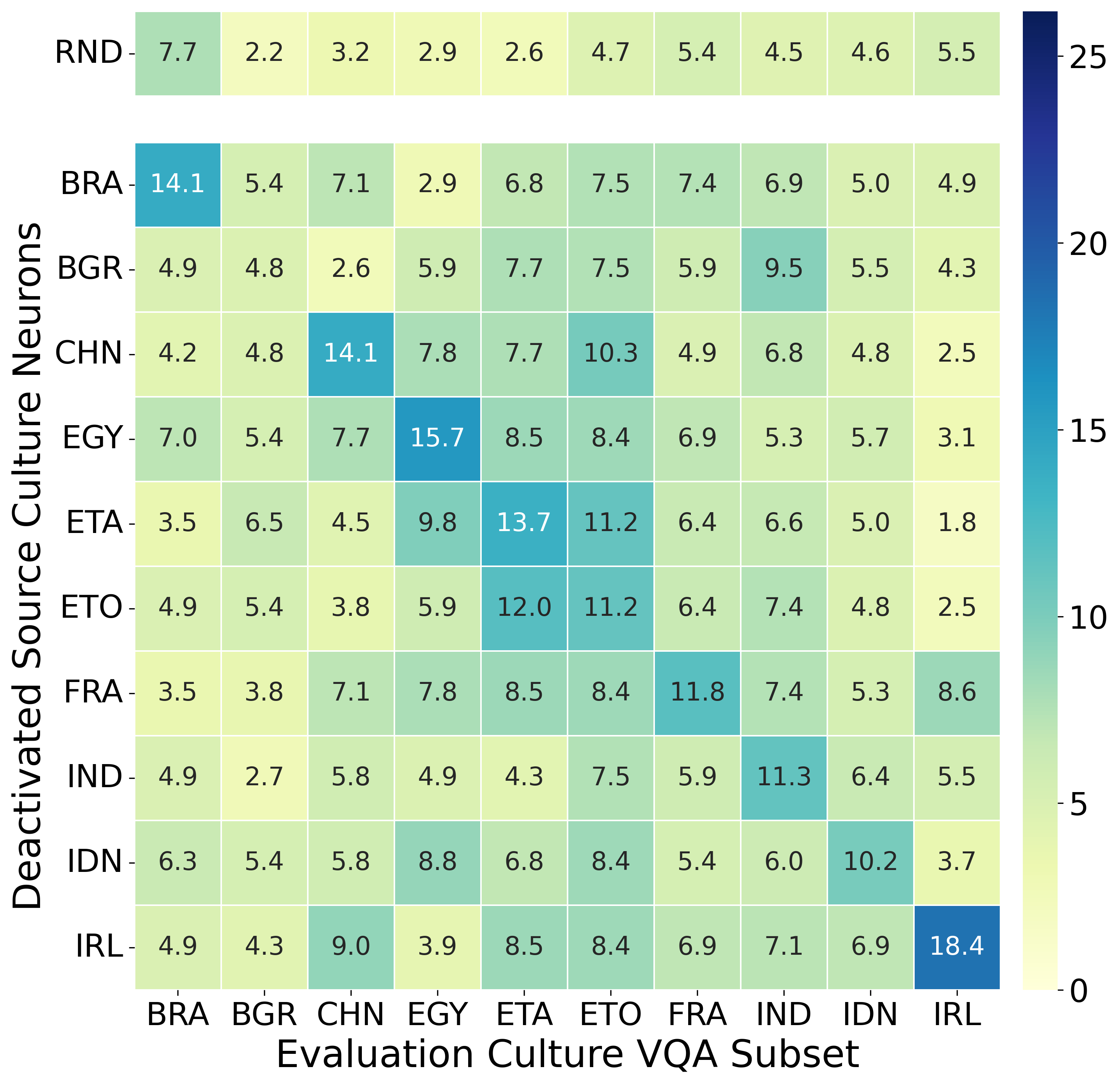}
        \caption{\textbf{MAD}}
    \end{subfigure}
    \hfill
    \begin{subfigure}[b]{0.24\textwidth}
        \centering
        \includegraphics[width=\textwidth]{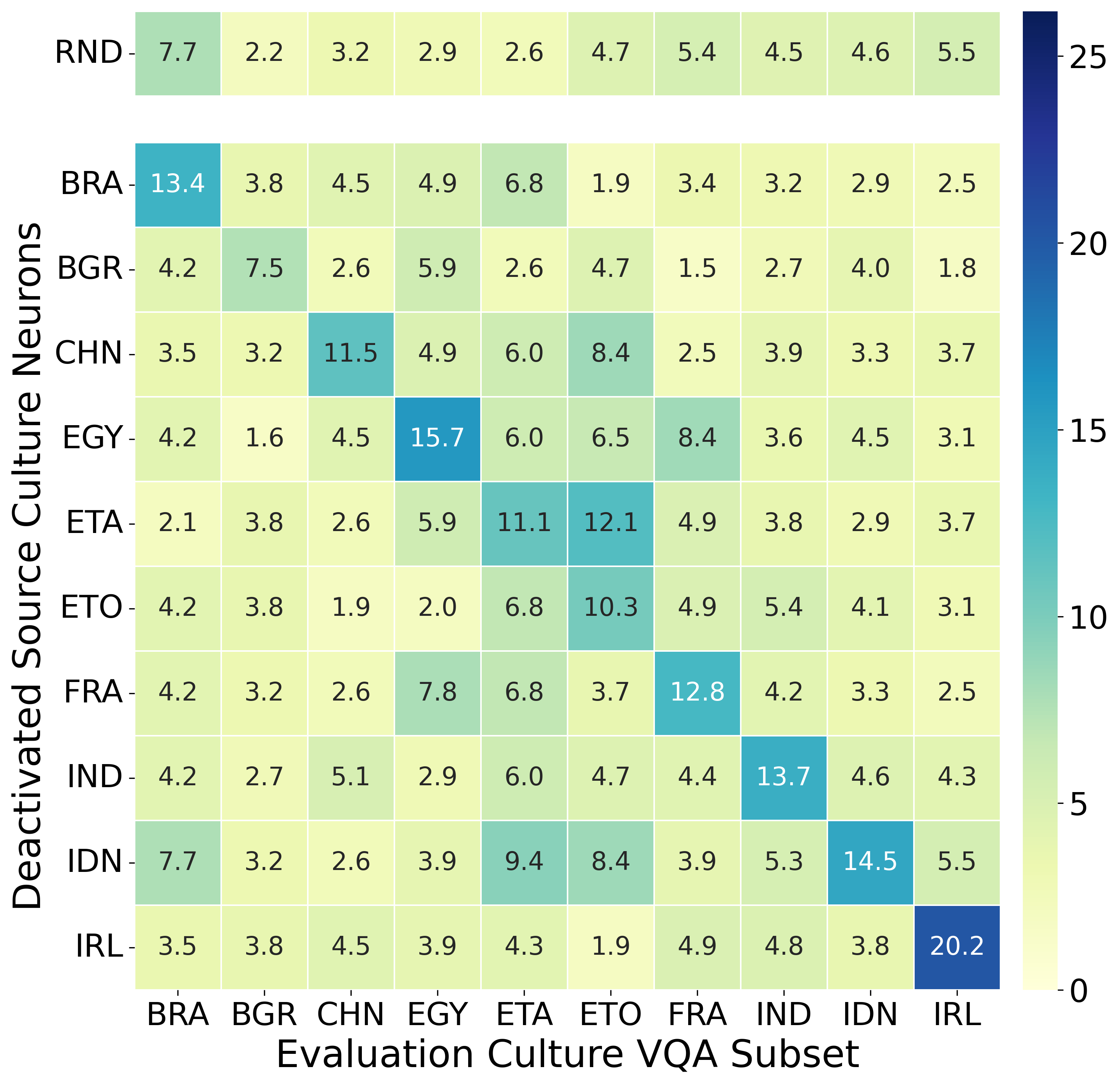}
        \caption{\textbf{ConAct}}
    \end{subfigure}

    \caption{\textbf{Accuracy change $\Delta$} (top) and \textbf{flip rate} heatmaps (bottom) on CVQA for different identification methods (Qwen2.5-VL-7B; showing the first ten evaluated cultures). On the y-axis we have the source culture for which neurons are identified and ablated and on the x-axis the culture used for evaluation. We report signed changes relative to the unablated full model. Diagonal cells show self-deactivation results. 
    }
    \label{fig:acc_qwen}
\end{figure*}

\subsection{Baseline Model Performance on CVQA}
\label{subsec:unmasked}

We first assess the unablated full model performance on CVQA, shown in Figure~\ref{fig:train_box}. All three VLMs exhibit substantial variation in performance across cultures. Qwen2.5-VL-7B achieves the highest median accuracy ($\approx0.60$), while Pangea-7B and LLaVA-v1.6-Mistral-7B reach around $0.50$. Importantly, identification and evaluation splits yield similar performance, suggesting that subsequent ablation results are not confounded by train--test mismatch. 
Overall, the models show uneven cultural competence but stable baselines, providing a reliable reference point for neuron ablations.

\subsection{Culture-Sensitive Neuron Ablation} 
\label{subsec:ablation}

Table~\ref{table:main_results} reports accuracy change ($\Delta$) and flip rates when deactivating neurons selected by each identification method (\S\ref{subsec:identification}). We analyze two evaluation settings as defined in \S~\ref{subsec:measuring_cultural_sensitivity}: \emph{self-deactivation} (masking neurons identified for the same culture as the evaluation set) and \emph{cross-deactivation} (masking neurons identified for a different culture).

For Qwen2.5-VL-7B and Pangea-7B, ConAct yields the largest self-deactivation drops in accuracy (Qwen: $-5.52$; Pangea: $-4.33$) paired with small cross-deactivation changes ($<1$), showing that the selected neurons are both important and relatively specific to their source culture. The associated flip rate gaps are likewise large and positive, indicating that predictions change substantially only within the target culture. By contrast, LAP and MAD often produce broader off-diagonal interference, capturing neurons linked to shared or generic multimodal cues rather than culture-specific signals. Occasionally, LAP even improves performance upon masking. Such gains can arise in ablation studies when removed units encode spurious or overly dominant features, effectively acting as a form of pruning \cite{ali2025detecting}.
For LLaVA-v1.6-Mistral-7B, LAPE induces the strongest self-deactivation drop ($-4.43$) but also larger cross-cultural spillover. ConAct and MAD yield neurons that show more cultural selectivity yet smaller in effect magnitude.

\paragraph{Culture-Specific patterns.} 
Figure~\ref{fig:acc_qwen} visualizes accuracy changes ($\Delta$) and flip rates for Qwen2.5-VL-7B of the first ten cultures.
We find that LAP shows broad column-shaped reductions (large off-diagonals), pointing to less specific features; LAPE reveals fairly little selectivity. On the contrary, both MAD and ConAct produce sharp diagonal degradations with limited but non-negligible off-diagonal changes, while ConAct achieves the cleanest separation between self and cross conditions.
The results for MAD and ConAct evidence strong mapping between masked neuron sets and cultures.
Additionally, some geographically linked cultures show correlated effects. For instance, deactivating EGY-neurons yields significant impact on ETO, another African culture group.
Interestingly, for some cultures we observe small accuracy gains even when ablating neuron sets identified from other cultures (e.g., BGR, ETA). A plausible explanation is that certain selected units capture spurious cues that hurt generalization on those subsets; removing them can therefore resemble pruning rather than culture-targeted disruption \citep{ali2025detecting}.

\subsection{Distribution Patterns across Layers}
\label{subsec:layer}
\begin{figure}[t!]
    \centering
    \begin{subfigure}{\columnwidth}
        \centering
        \includegraphics[width=\columnwidth]{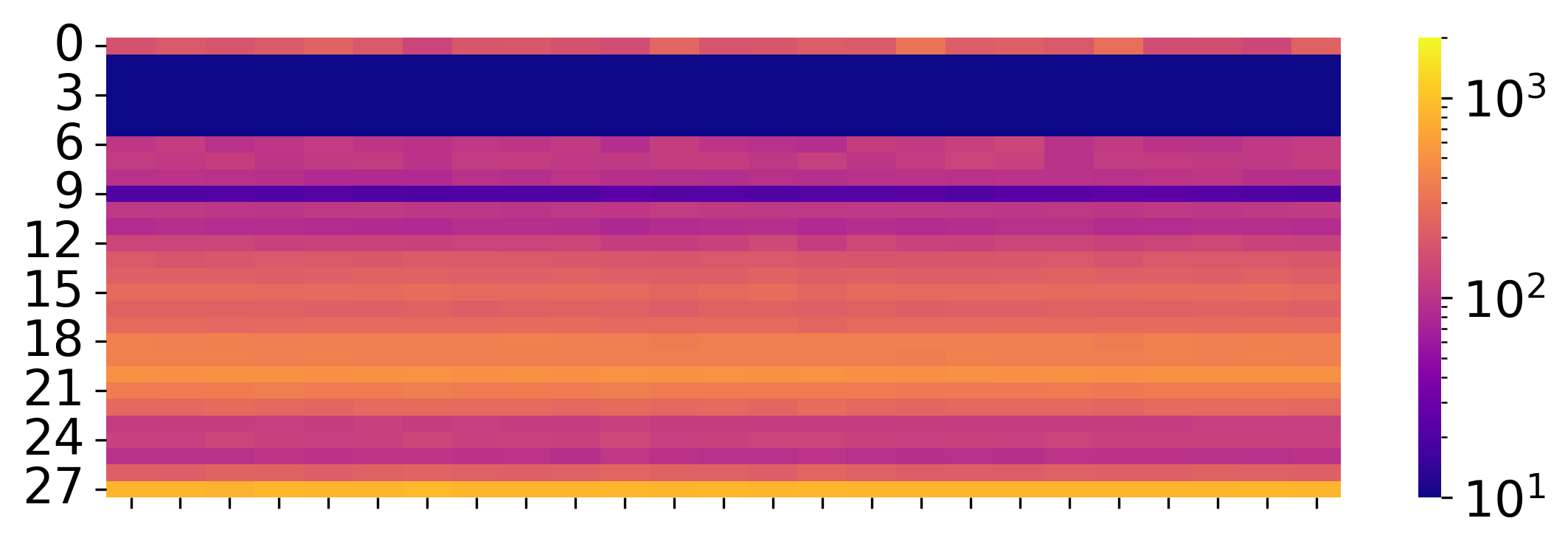}
        \caption{\textbf{LAP}}
    \end{subfigure}

    \begin{subfigure}{\columnwidth}
        \centering
        \includegraphics[width=\columnwidth]{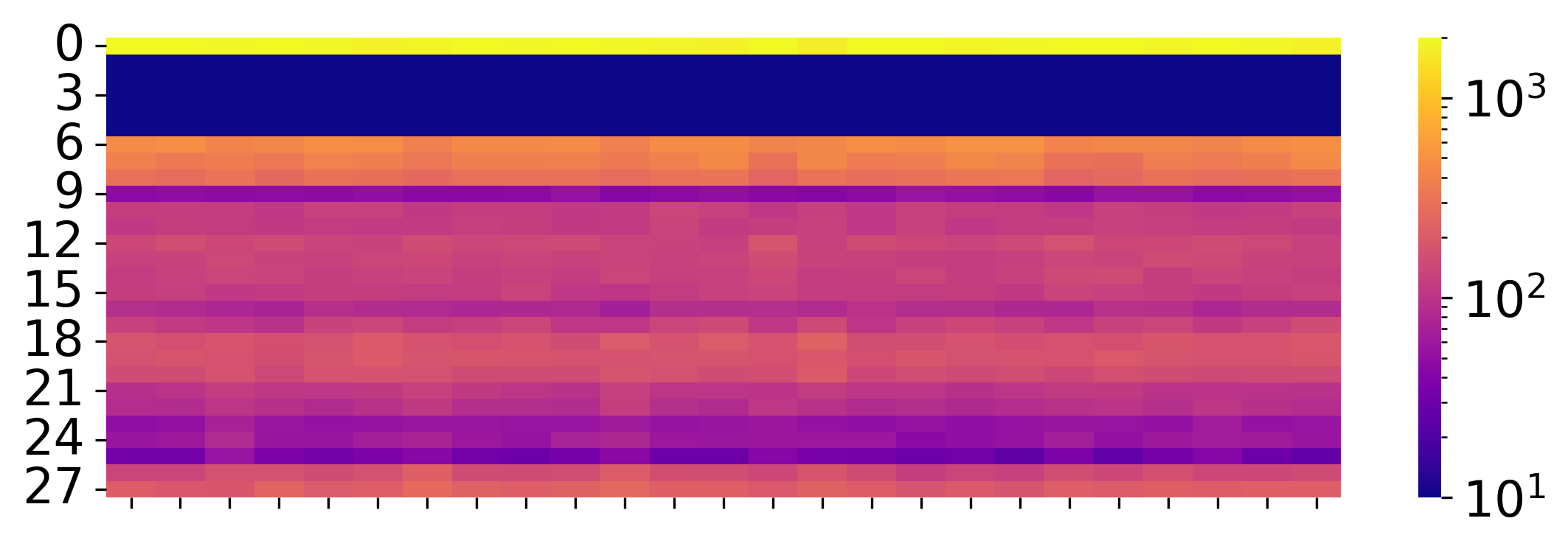}
        \caption{\textbf{LAPE}}
    \end{subfigure}

    \begin{subfigure}{\columnwidth}
        \centering
        \includegraphics[width=\columnwidth]{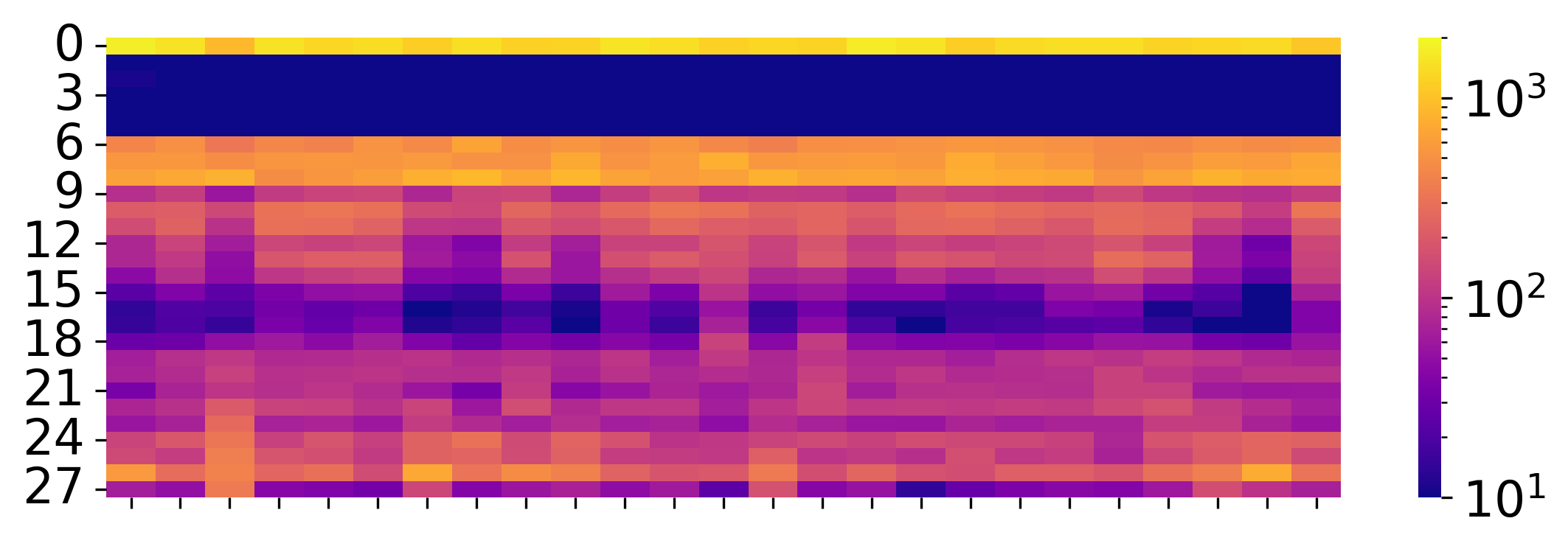}
        \caption{\textbf{MAD}}
    \end{subfigure}

    \begin{subfigure}{\columnwidth}
        \centering
        \includegraphics[width=\columnwidth]{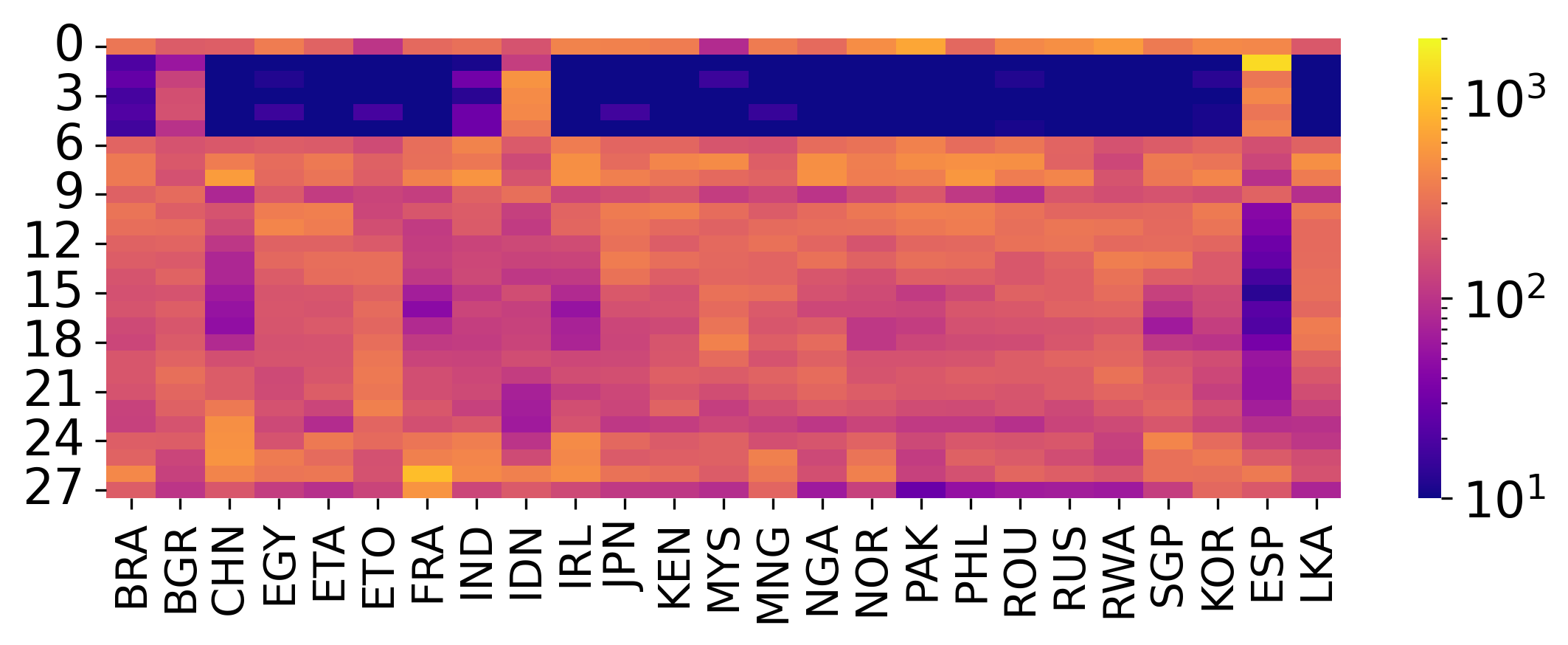}
        \caption{\textbf{ConAct}}
    \end{subfigure}

    \caption{\textbf{Layer-wise counts of identified neurons} by different methods (Qwen2.5-VL-7B; log-scaled color).}
    \label{fig:layer_heatmap_qwen}
\end{figure}

Figure~\ref{fig:layer_heatmap_qwen} shows the layer-wise distribution of culture-sensitive neurons identified in Qwen2.5-VL-7B (28-layer decoder). Understanding where such neurons concentrate within the network can offer clues about how culture-related information is integrated e.g.,\ whether it is handled early, during basic feature fusion, or later, during high-level reasoning. Moreover, comparing distributions across identification methods reveals whether different methods capture similar or distinct functional subspaces, while cross-cultural differences can hint at culture-specific processing pathways.

We observe that culture-sensitive neurons generally cluster in the first layer (layer 0) and the early-mid layers (6--8), with relatively sparse presence in deeper blocks. Interestingly, MAD tends to bypass the central layers (15--18), whereas ConAct identifies neurons more evenly across mid-to-late layers. ConAct also shows culture-specific deviations, for example, in BGR and IDN, layers 6--8 contain a higher proportion of selected neurons than in other cultures. These patterns suggest that both the choice of method and culture influence which layers of the model are most engaged in culturally grounded processing.

\subsection{Effect of Ablation on Model Behavior}
Figure~\ref{fig:example} highlights two key observations about how ablation disrupts model behavior. 
First, we do not observe widespread degradation of format compliance under decoder-level masking: masked generations typically remain compatible with the multiple-choice response format (Appendix~\ref{appendix:normalize}), suggesting that the intervention does not generally collapse task framing at the sparsity levels we study.
Second, we find that different identification methods perturb cultural knowledge in distinct ways: RND yields only small changes, suggesting that arbitrary neurons are rarely detrimental for culture-specific performance. In contrast, LAPE and ConAct push the model to different incorrect but plausible options. This suggests that the ablated neurons induce selective culture degradation.

Overall, our analyses reveal several consistent patterns across 25 cultural groups and three model architectures: (1) A select subset of decoder neurons exhibit clear culture-sensitive activation patterns, suggesting that cultural knowledge is at least in part encoded locally. %
(2) These neurons play an important role in culturally grounded processing: their removal selectively degrades performance on the corresponding culture while largely preserving performance elsewhere. 
(3) Culture-sensitive neurons are not uniformly distributed but cluster in early to mid decoder layers (with some model- and culture-specific variation).
(4) Among all identification methods, ConAct most effectively isolates such neurons.%

\section{Conclusion}
\label{sec:conclusion}

This study provides empirical evidence for the existence of culture-sensitive neurons in VLMs by showing inference-time ablations of targeted subsets of neurons that selectively disrupt VLMs' culture-specific performance. We introduce a margin-based selector (ConAct) that allows for more precise identification of culture-sensitive neurons. 
Among the identification methods we compare, ConAct identifies neurons whose ablation yields the largest self-deactivation drops with minimal cross-deactivation spillover on Qwen2.5-VL-7B and Pangea-7B, while LLaVA-v1.6-Mistral-7B shows resistance to specific targeting. 
Layer-wise analyses further show that the identified neurons concentrate in specific decoder regions, with distributions that vary by model and selector.
Overall, our findings suggest that small, targeted neuron suites can serve as a diagnostic handle for probing culturally grounded behavior in VLMs, and that margin-based selection can better isolate units whose ablation yields large self--cross gaps in some architectures. Future work could extend the search beyond decoder MLPs and pair identification with activation steering.

\newpage

\section*{Limitations}
\label{subsec:limitations}
\paragraph{Defining ``culture''.}
We use CVQA's country--language taxonomy and, for fairness to multilingual models, only the English-translated prompts to better decouple language skill from cultural recognition. This choice makes the construct closer to visual cultural knowledge than to culture-as-language-practice \cite{kramsch2014language}. For multilingual models, it remains unknown whether our observations would still emerge, which we leave for future work.

\paragraph{Model components.}
Our analysis is restricted to decoder MLP neurons and does not cover attention heads, vision encoders, or alignment modules, which may also encode culture-sensitive behavior. We rely on activation-frequency summaries rather than more fine-grained temporal or token-level dynamics, and we fix hyperparameters for neuron selection based on computational budget.

\section*{Ethical Considerations}
This study aims to improve transparency and fairness in multimodal models by examining culture-sensitive neurons. All experiments are conducted on a publicly available dataset (CVQA), and no new human subject data or personally identifiable information is used.

A potential ethical concern lies in the definition of ``culture.'' For experimental feasibility, we adopt CVQA's taxonomy of country--language pairs and, in some cases, group multiple pairs that share a common country or language tag. Such grouping is a dataset-driven simplification and does not reflect the diversity, fluidity, or internal variation within cultural communities. Our results should not be interpreted as essentializing or stereotyping real-world cultures but rather as insights into how models respond to the categories provided by the benchmark.

The methods presented are intended for diagnostic use only. While they can help reveal and quantify cultural disparities in model behavior, they are not in themselves fairness interventions. Misuse of these methods to draw normative claims about communities would be harmful and contrary to the goals of this work. We encourage future studies to incorporate broader and more inclusive datasets when assessing and mitigating cultural bias in multimodal systems.

\section*{Acknowledgments}
We thank Simon King, Korin Richmond, and Catherine Lai at the University of Edinburgh for their constant support during the course of the project. Special thanks to Jinzuomu Zhong for providing help on computational resources.

\bibliography{custom}

\appendix

\section{Reproducibility}
\label{appendix:reproduce}
\subsection{Models and Sources}
\begin{table}[h]
\centering
\resizebox{\columnwidth}{!}{%
\begin{tabular}{@{}ll@{}}
\toprule
\textbf{Models}        & \textbf{Sources}                   \\ \midrule
LLaVA-v1.6-Mistral-7B   & \url{https://huggingface.co/llava-hf/LLaVA-v1.6-Mistral-7B-hf}   \\
Pangea-7B   & \url{https://huggingface.co/neulab/Pangea-7B}  \\
Qwen2.5-VL-7B     & \url{https://huggingface.co/Qwen/Qwen2.5-VL-7B-Instruct}                  \\ \bottomrule
\end{tabular}%
}
\caption{
Sources of the evaluated models.
}
\label{table:models}
\end{table}

\subsection{Culture Grouping of CVQA}
\label{subsec:grouping}
The CVQA benchmark comprises 39 country--language pairs, several of which share the same country or language tags. To study potential grouping effects, we construct three aggregated culture sets that pool pairs with a shared attribute: India--all (IND) (all pairs tagged with country ``India''), Indonesia--all (IDN) (all pairs tagged with country ``Indonesia''), and all--Spanish (ESP) (all pairs whose language is Spanish). Table~\ref{table:grouping_full} reports the mapping from individual pairs to each aggregate and the number of questions per subset in the identification and evaluation splits.

\begin{table}[ht!]
\centering
\resizebox{\columnwidth}{!}{%
\begin{tabular}{@{}lccc@{}}
\toprule
CVQA Pairs            & Grouped Cultures     & \# Qs (I) & \# Qs (E) \\ \midrule
Brazil--Portuguese     & BRA                  & 142       & 142       \\
Bulgaria--Bulgarian    & BGR                  & 185       & 186       \\
China--Chinese         & CHN                  & 155       & 156       \\
Egypt--Egyptian Arabic & EGY                  & 101       & 102       \\
Ethiopia--Amharic      & ETA                  & 117       & 117       \\
Ethiopia--Oromo        & ETO                  & 107       & 107       \\
France--Breton         & FRA                  & 202       & 203       \\ \midrule
India--Bengali         & \multirow{6}{*}{IND} & 143       & 143       \\
India--Hindi           &                      & 100       & 101       \\
India--Marathi         &                      & 101       & 101       \\
India--Tamil           &                      & 107       & 107       \\
India--Telugu          &                      & 100       & 100       \\
India--Urdu            &                      & 110       & 110       \\ \midrule
Indonesia--Indonesian  & \multirow{4}{*}{IDN} & 206       & 206       \\
Indonesia--Javanese    &                      & 148       & 149       \\
Indonesia--Minangkabau &                      & 125       & 126       \\
Indonesia--Sundanese   &                      & 100       & 100       \\ \midrule
Ireland--Irish         & IRL                  & 163       & 163       \\
Japan--Japanese        & JPN                  & 101       & 102       \\
Kenya--Swahili         & KEN                  & 136       & 137       \\
Malaysia--Malay        & MYS                  & 157       & 158       \\
Mongolia--Mongolian    & MNG                  & 156       & 156       \\
Nigeria--Igbo          & NGA                  & 100       & 100       \\
Norway--Norwegian      & NOR                  & 146       & 150       \\
Pakistan--Urdu         & PAK                  & 108       & 108       \\
Philippines--Filipino  & PHL                  & 101       & 102       \\
Romania--Romanian      & ROU                  & 151       & 151       \\
Russia--Russian        & RUS                  & 100       & 100       \\
Rwanda--Kinyarwanda    & RWA                  & 117       & 118       \\
Singapore--Chinese     & SGP                  & 106       & 106       \\
South Korea--Korean    & KOR                  & 145       & 145       \\ \midrule
Argentina--Spanish     & \multirow{7}{*}{ESP} & 132       & 133       \\
Chile--Spanish         &                      & 117       & 117       \\
Colombia--Spanish      &                      & 120       & 121       \\
Ecuador--Spanish       &                      & 181       & 181       \\
Mexico--Spanish        &                      & 161       & 162       \\
Spain--Spanish         &                      & 159       & 159       \\
Uruguay--Spanish       &                      & 157       & 158       \\ \midrule
Sri Lanka--Sinhala     & LKA                  & 112       & 113       \\ \midrule
Total                 &                      & 5178      & 5196      \\ \bottomrule
\end{tabular}%
}
\caption{\textbf{Culture subsets and VQA statistics}. CVQA country--language pairs with ``Spanish'' language tag or [``India'', ``Indonesia''] country tag are assigned to one of the aggregated cultures, and other pairs remain stand-alone. \textbf{\# Qs (I)} denotes the number of questions used for activation recording and neuron identification, while \textbf{\# Qs (E)} denotes the number of questions used for masked generation and evaluation.}
\label{table:grouping_full}
\end{table}

\newpage

\subsection{Prompt Template for Multiple-Choice VQA}
\label{appendix:prompt}
\begin{lstlisting}[style=promptbox, caption={Prompt template used for VQA generation}, label={lst:mcq-prompt}]
Answer the following multiple-choice question based on the image.

Question: 
{question}

Options:
{option 1}
{option 2}
{option 3}
{option 4}

Your response must be ONLY the text of the correct option from the list above, and nothing else.
\end{lstlisting}

\subsection{Answer Normalization Process}
\label{appendix:normalize}

To ensure reliable evaluation of model predictions in the multiple-choice setting, we implemented a normalization procedure to mitigate inconsistencies in the format and phrasing of generated outputs.

First, the prediction string is converted to lowercase and standardized by collapsing all whitespace into single spaces and trimming leading and trailing spaces. This step minimizes mismatches caused by case sensitivity or formatting irregularities. Each answer option is similarly normalized to lowercase. The algorithm then searches for whole-word matches of each choice within the normalized prediction using word-boundary matching to prevent false positives.  
Because LLMs are known to exhibit label bias in multiple-choice answering settings \cite{zheng2024large, zhao-etal-2024-measuring}, we require the model to output the full content of the chosen option rather than its label (e.g., ``A'', ``B'').

Although the prompt explicitly instructs the model to generate a single answer (Appendix~\ref{appendix:prompt}), instruction-tuned language models may still produce extended reasoning, which makes simple substring matching insufficient. Therefore, we applied a heuristic when multiple choices appear in the output: the last-mentioned choice is treated as the model's final decision. This heuristic reflects the common generation pattern where models deliberate over several options before declaring a final answer (e.g., ``Option A is plausible, but B is incorrect, so the answer is C''). If multiple choices appear in the output, we treat the last-mentioned choice as the model's final decision, reflecting common generations that deliberate over several options before committing to an answer.
If no choice can be confidently identified using whole-word matching and the last-mentioned rule above, we count the prediction as incorrect for accuracy.

This two-stage normalization and extraction process improves the evaluation's robustness to varied model output styles while prioritizing the most plausible interpretation of the model's intended final answer.

\section{Layer-Wise Neuron Distribution}
\label{sec:app_layers}
\begin{figure}[ht!]
    \centering
    \begin{subfigure}{\columnwidth}
        \centering
        \includegraphics[width=\columnwidth]{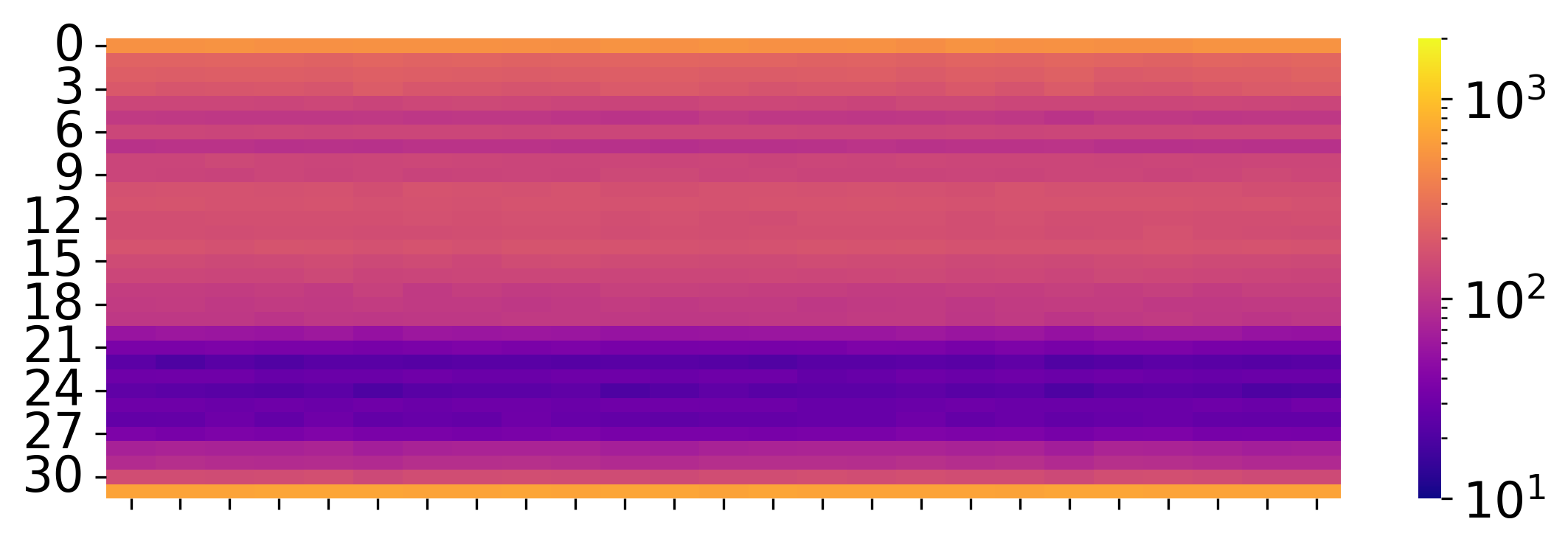}
        \caption{\textbf{LAP}}
    \end{subfigure}
        \hfill
    \begin{subfigure}{\columnwidth}
        \centering
        \includegraphics[width=\columnwidth]{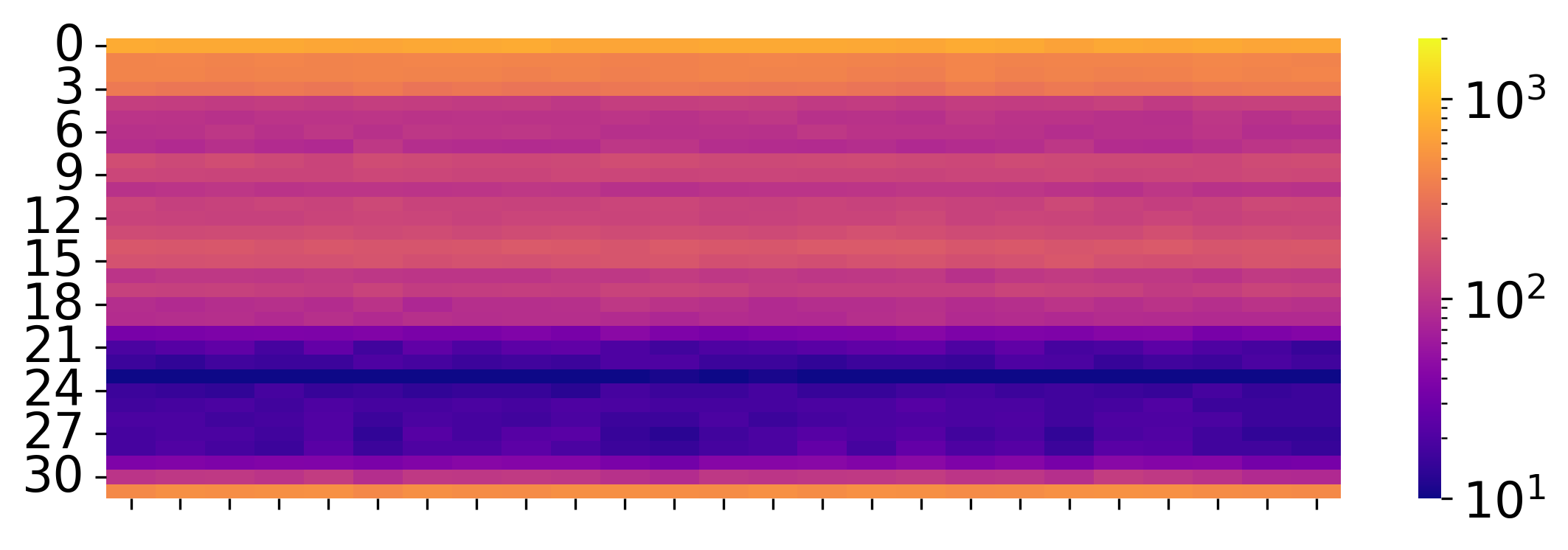}
        \caption{\textbf{LAPE}}
    \end{subfigure}
    \begin{subfigure}{\columnwidth}
        \centering
        \includegraphics[width=\columnwidth]{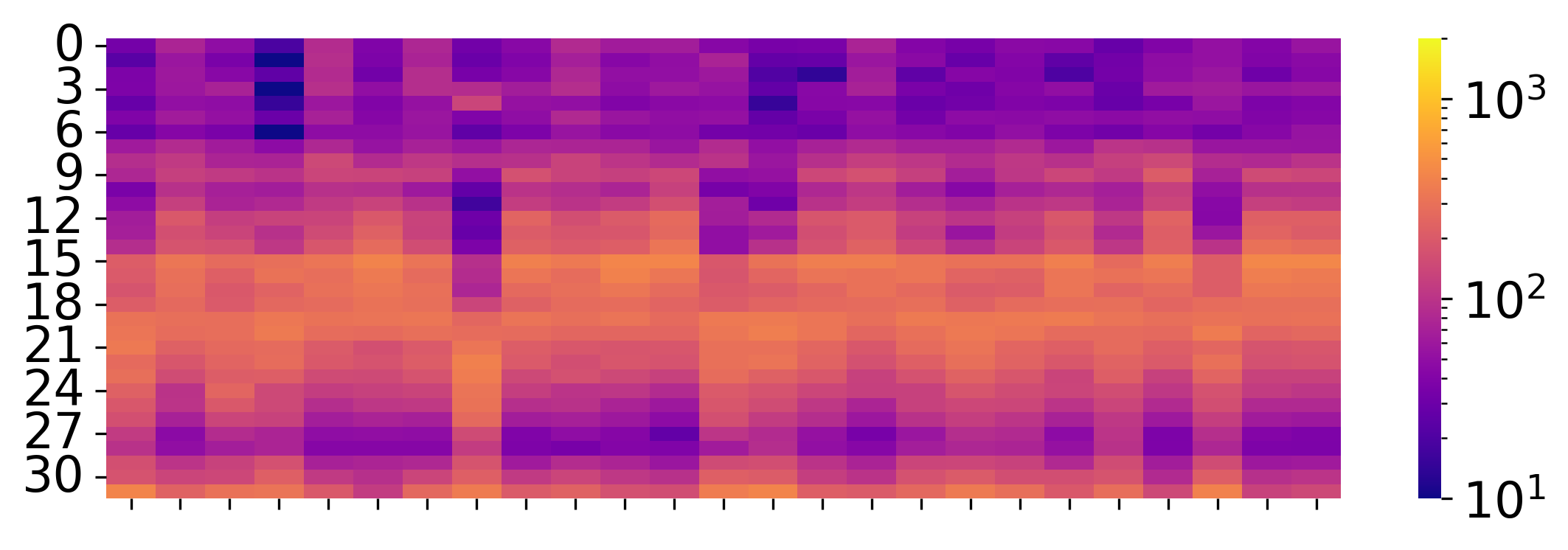}
        \caption{\textbf{MAD}}
    \end{subfigure}
    \hfill
    \begin{subfigure}{\columnwidth}
        \centering
        \includegraphics[width=\columnwidth]{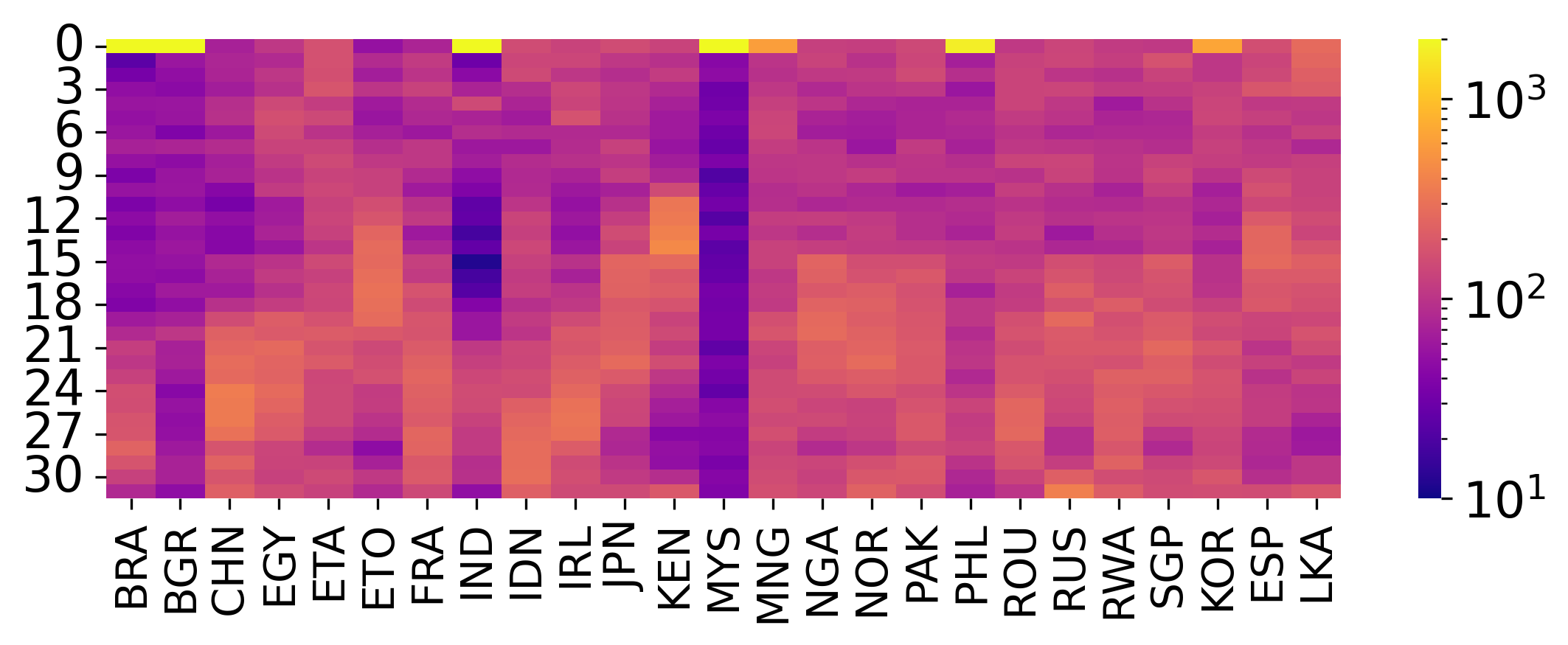}
        \caption{\textbf{ConAct}}
    \end{subfigure}

    \caption{\textbf{Layer-wise counts of identified neurons (LLaVA-v1.6-Mistral-7B).}}
    \label{fig:layer_heatmap_llava}
\end{figure}

\begin{figure}[ht!]
    \centering
    \begin{subfigure}{\columnwidth}
        \centering
        \includegraphics[width=\columnwidth]{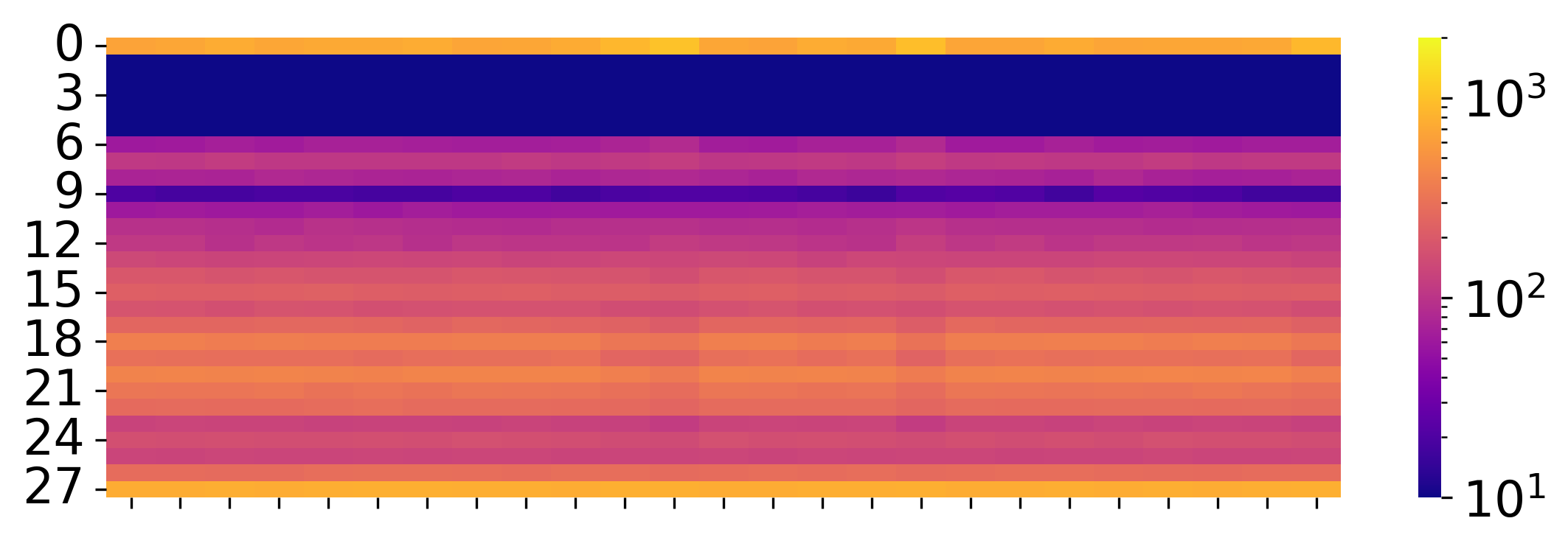}
        \caption{\textbf{LAP}}
    \end{subfigure}
        \hfill
    \begin{subfigure}{\columnwidth}
        \centering
        \includegraphics[width=\columnwidth]{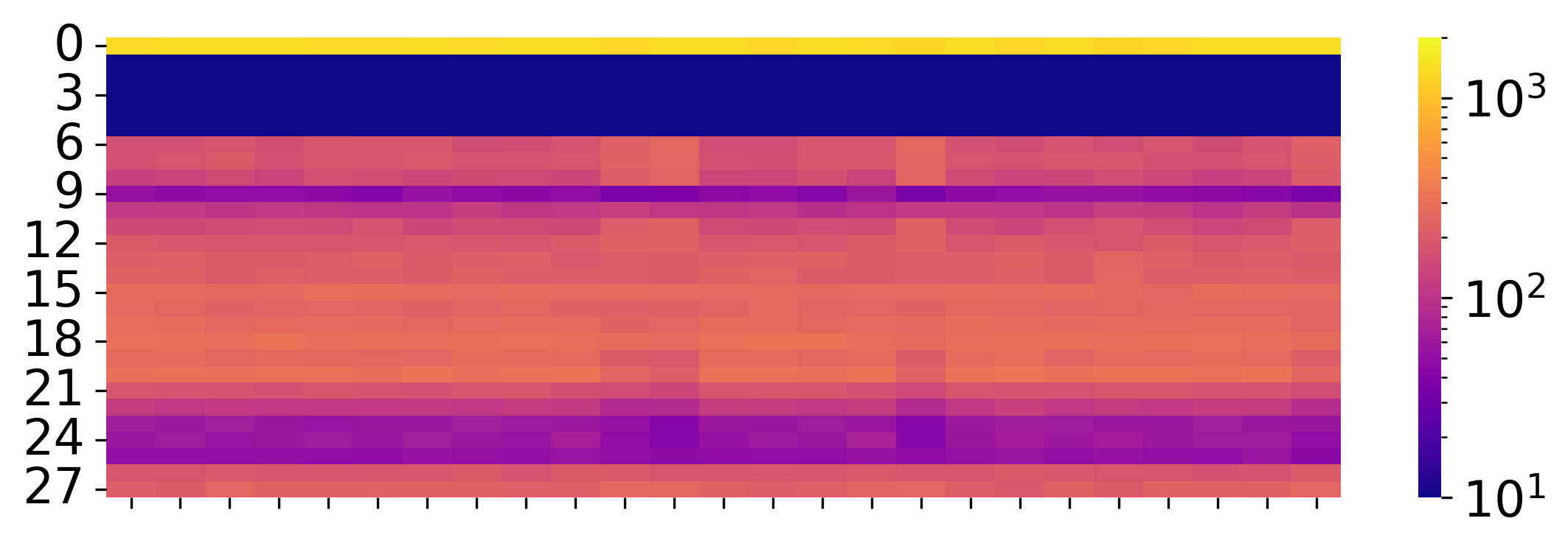}
        \caption{\textbf{LAPE}}
    \end{subfigure}
    \begin{subfigure}{\columnwidth}
        \centering
        \includegraphics[width=\columnwidth]{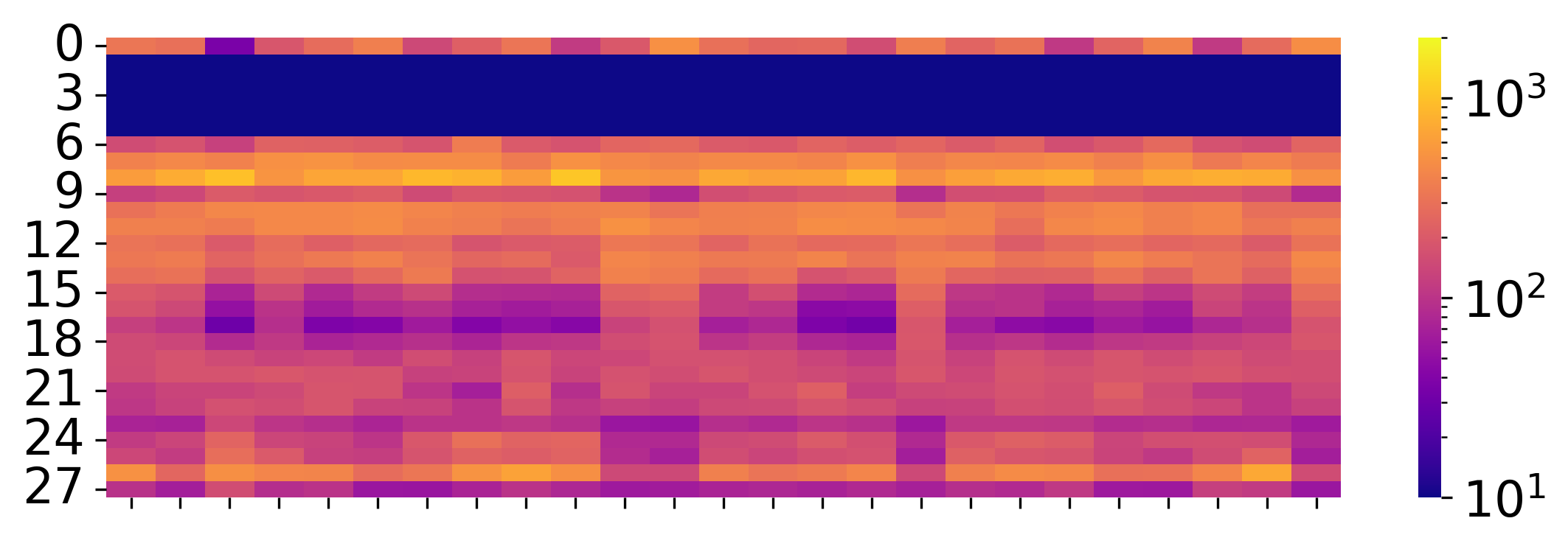}
        \caption{\textbf{MAD}}
    \end{subfigure}
    \hfill
    \begin{subfigure}{\columnwidth}
        \centering
        \includegraphics[width=\columnwidth]{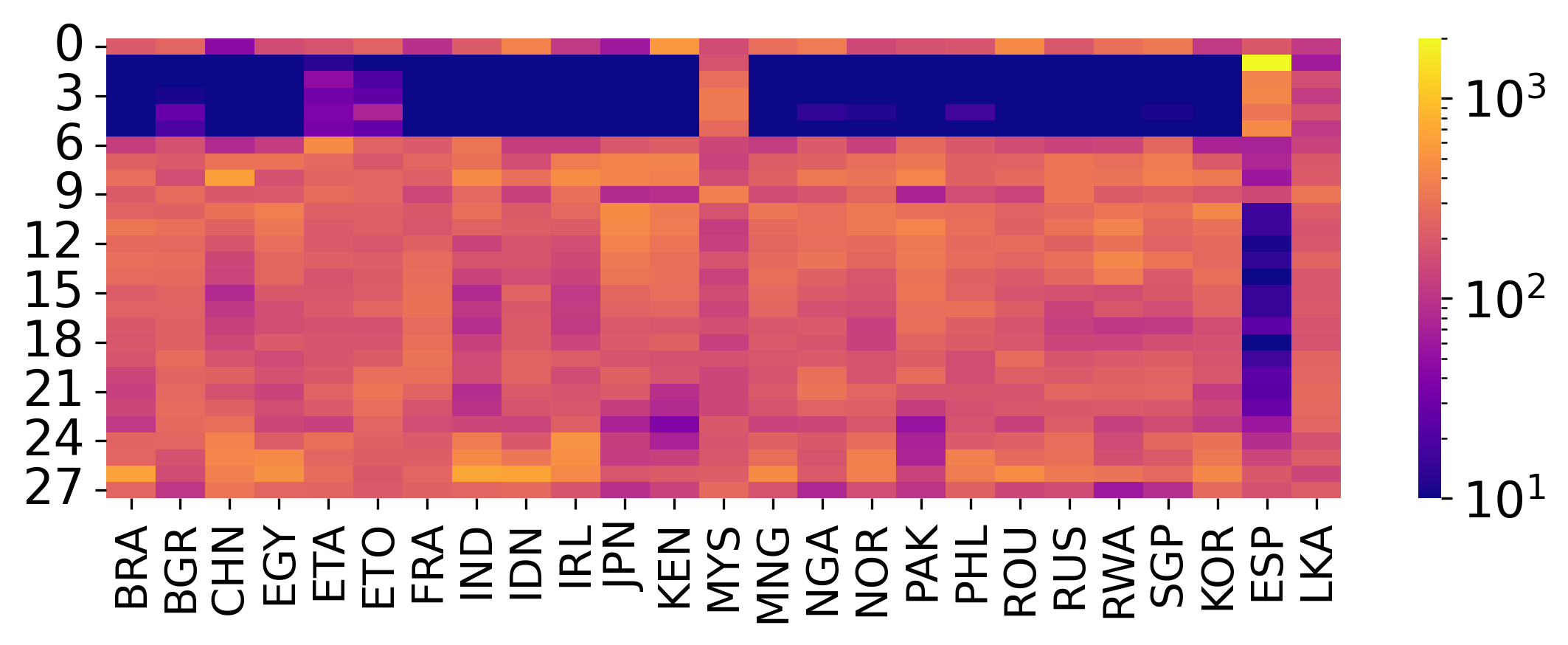}
        \caption{\textbf{ConAct}}
    \end{subfigure}

    \caption{\textbf{Layer-wise counts of identified neurons (Pangea-7B).}}
    \label{fig:layer_heatmap_pangea}
\end{figure}

We report the layer-wise counts of culture-sensitive neurons identified in LLaVA-v1.6-Mistral-7B (Figure~\ref{fig:layer_heatmap_llava}) and Pangea-7B (Figure~\ref{fig:layer_heatmap_pangea}), aggregated over all selected cultures. To improve visual comparability across layers, all heatmaps use a logarithmic color scale, which compresses extremely large counts in dominant layers while expanding the dynamic range for smaller counts elsewhere.

\label{subsubsec:layers_llava}

\paragraph{LLaVA-v1.6-Mistral-7B}
Culture-sensitive neuron distributions in LLaVA-v1.6-Mistral-7B vary more across identification methods than in Qwen2.5-VL-7B and Pangea-7B. LAP and LAPE concentrate strongly in the earliest and latest layers, while still covering a broad early-to-mid region (\mbox{1--20}). In contrast, both show a pronounced low-activation ``band'' around \mbox{21--27}, indicating limited selection in those layers. MAD primarily concentrates in mid-to-late layers (\mbox{15--24}). ConAct is generally sparser than MAD and selects relatively more neurons in early layers (\mbox{0--6}), suggesting a different locus of culture-selective evidence under the margin-based criterion.

\paragraph{Pangea-7B}

The layer-wise distribution for Pangea-7B closely mirrors Qwen2.5-VL-7B, consistent with the fact that Pangea-7B's language component is built on a Qwen2-7B-Instruct backbone \cite{yang2024qwen2technicalreport, yue2025pangea}, a direct predecessor of Qwen2.5-7B-Instruct \cite{qwen2025qwen25technicalreport} used by Qwen2.5-VL-7B. Both models share a 28-layer decoder architecture. Similar to Qwen2.5-VL-7B, the aggregated all--Spanish culture group (ESP) exhibits a distinctive early-layer concentration (\mbox{0--5}) compared to other cultures, suggesting that some shared language-associated or region-associated cues are preferentially localized to early decoder MLPs in these backbones.

\section{Granularity of Culture Grouping}\label{sec:granularity}
We provide two finer-grained analyses that probe culture grouping at different granularities using Qwen2.5-VL-7B with ConAct-selected neurons: (1) within a single country across multiple languages (India), and (2) across multiple countries within a shared language (Spanish). 

\subsection{Within a Single Country Across Languages: India}

\begin{figure}[ht!]
    \centering
    \begin{subfigure}[t]{0.48\columnwidth}
        \centering
        \includegraphics[width=\linewidth]{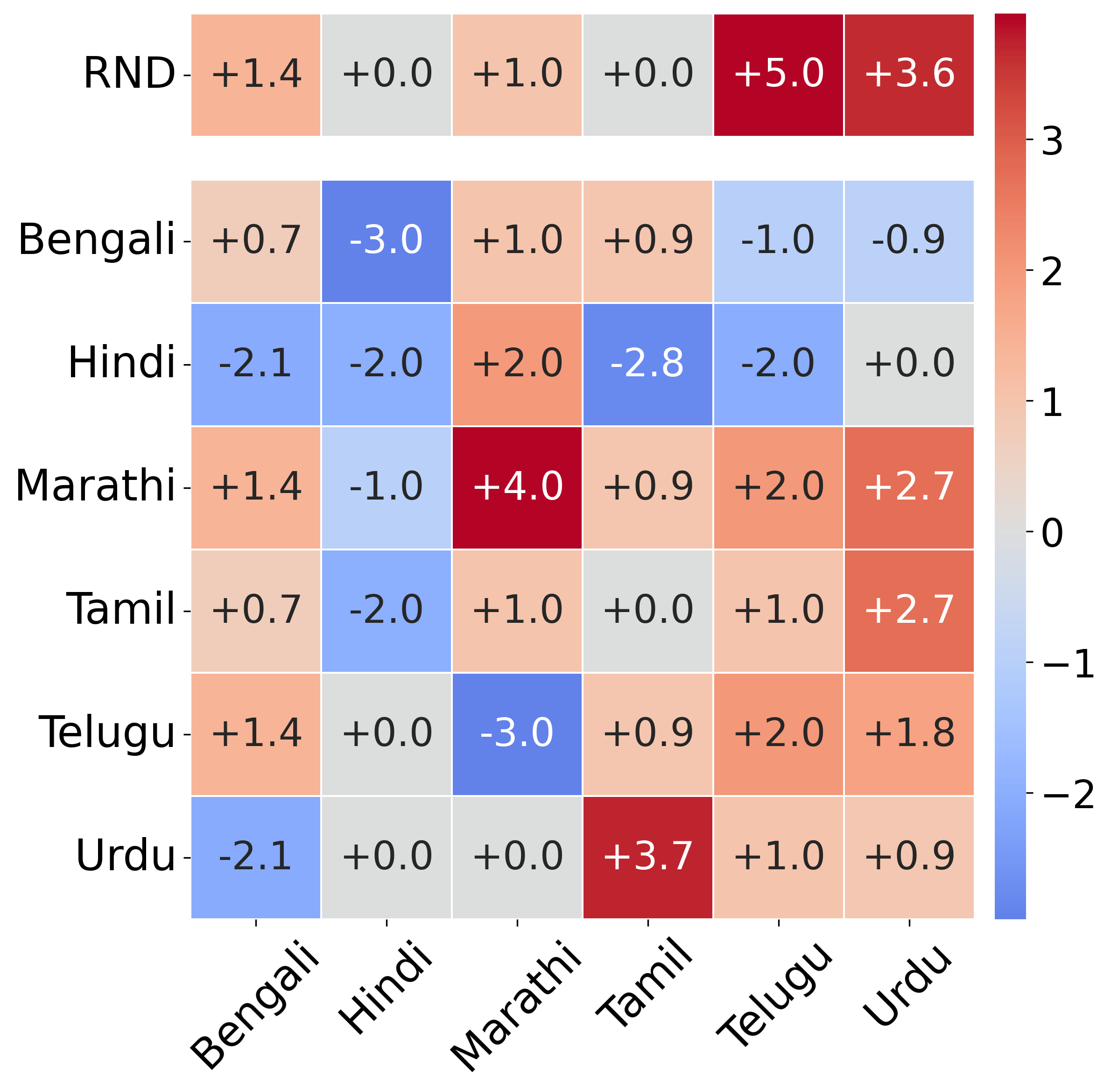}
        \caption{Accuracy changes $\Delta$ for India cultures.}
    \end{subfigure}
    \hfill
    \begin{subfigure}[t]{0.48\columnwidth}
        \centering
        \includegraphics[width=\linewidth]{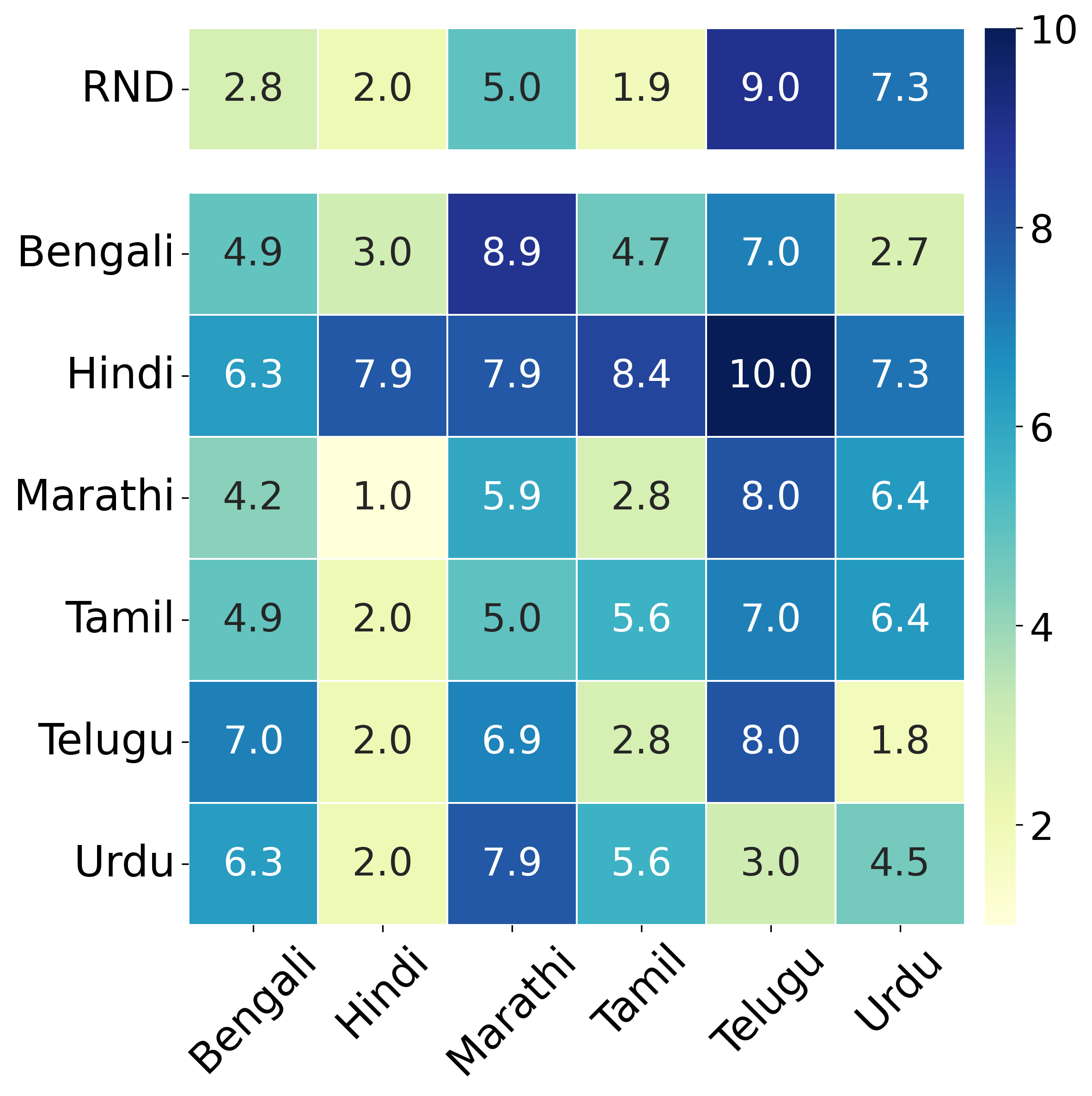}
        \caption{Flip rates for India cultures.}
    \end{subfigure}
    \caption{Within-country (India) cross-language ablations (ConAct-selected neurons on Qwen2.5-VL-7B).}
    
    \label{fig:intra-india_qwen}
\end{figure}

Figure~\ref{fig:intra-india_qwen} compares ablation effects across six India language-culture subsets (Bengali, Hindi, Marathi, Tamil, Telugu, Urdu). Overall self-deactivation effects on accuracy are modest, ranging from $-3.0$ to $+4.0$, but are more consistently negative for Hindi ($-2.0$), suggesting that the identified Hindi-specific neurons carry relatively stronger self-specific evidence.

The same-country interactions are mixed and occasionally beneficial (e.g., ablating Urdu-identified neurons yields $+3.7$ on Tamil; ablating Marathi-identified neurons yields $+4.0$ on Marathi), which is consistent with partial feature overlap and potential interference effects under shared national context. Such positive gains may also reflect pruning of detrimental or noisy features \citep{ali2025detecting}. Flip rates align with this interpretation: Hindi and Telugu exhibit higher self flip rates ($7.9$ and $8.0$), while off-diagonal flips are typically lower (about $1.5$--$6.8$).

\subsection{Across Countries Within a Shared Language: Spanish}

\begin{figure}[ht!]
    \centering
    \begin{subfigure}[t]{0.48\columnwidth}
        \centering
        \includegraphics[width=\columnwidth]{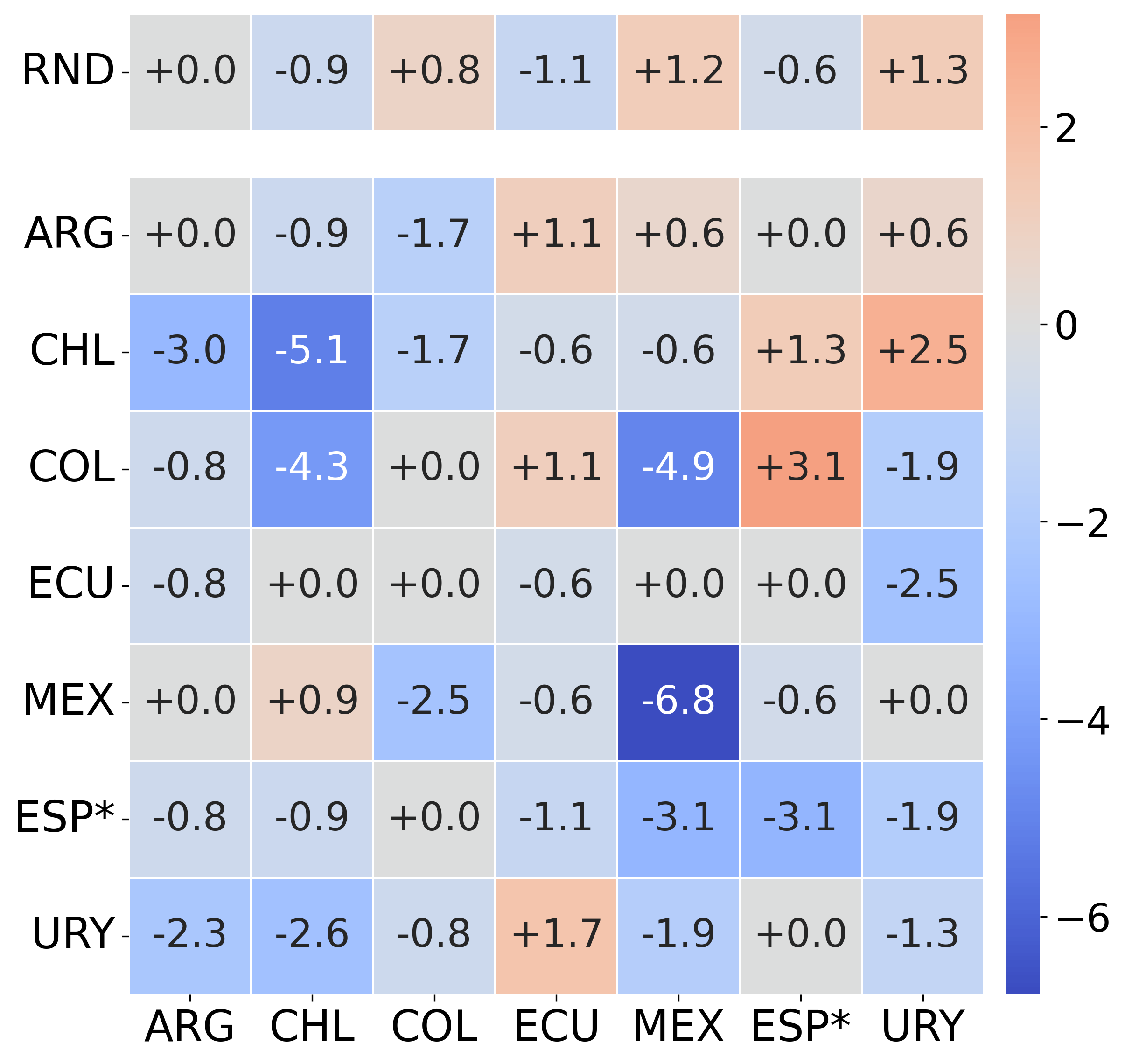}
        \caption{Accuracy changes $\Delta$ for Spanish-speaking cultures.}
    \end{subfigure}
    \hfill
    \begin{subfigure}[t]{0.48\columnwidth}
        \centering
        \includegraphics[width=\columnwidth]{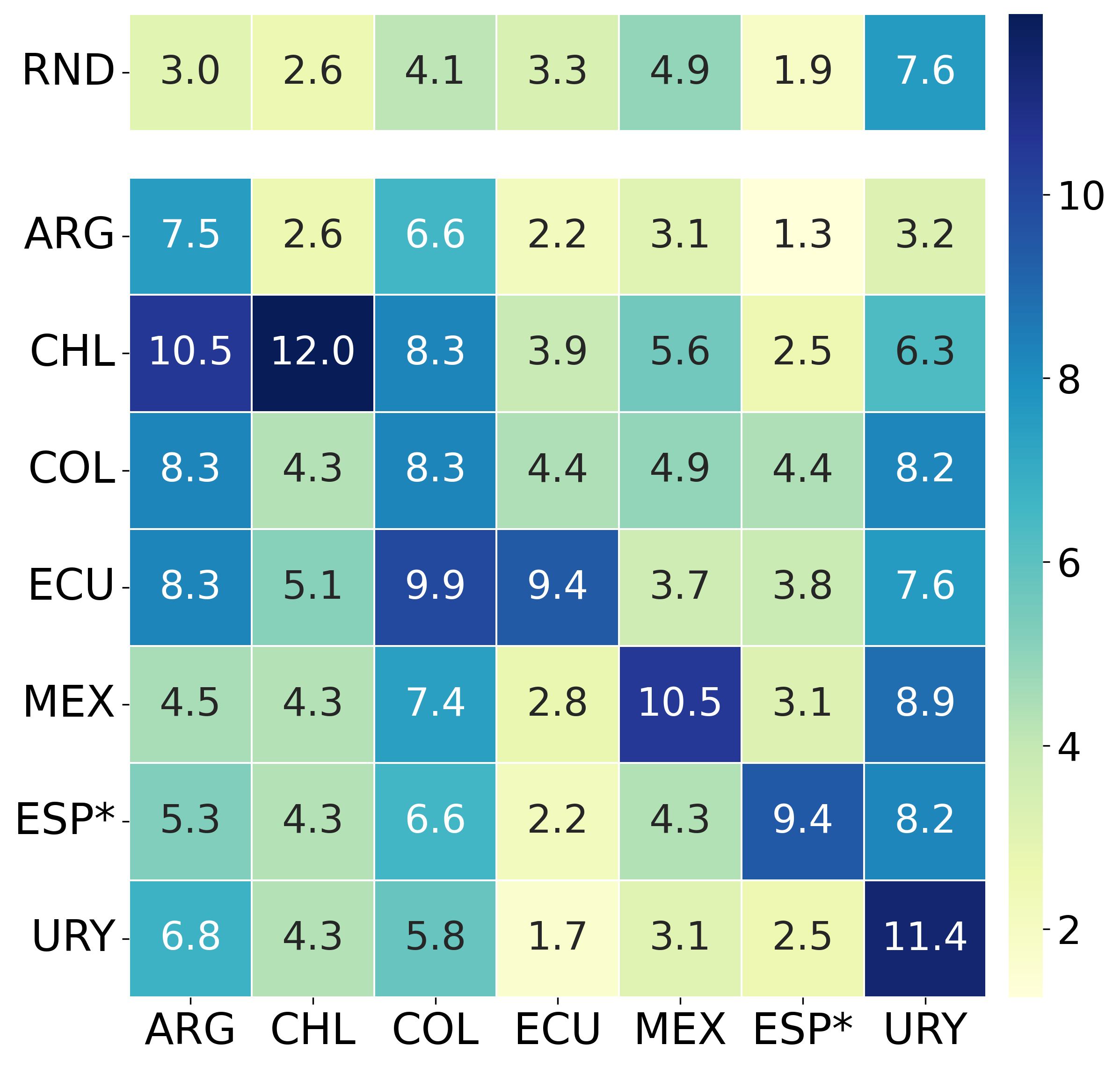}
        \caption{Flip rates for Spanish-speaking cultures.}
    \end{subfigure}
    \caption{Same-language (Spanish) cross-country ablations for Argentina, Chile, Colombia, Ecuador, Mexico, Spain, and Uruguay (ConAct-selected neurons on Qwen2.5-VL-7B). ``ESP*'' denotes Spain alone, distinct from ``ESP'' used elsewhere to denote the aggregated all--Spanish group.}
    \label{fig:intra-spanish-qwen}
\end{figure}

Figure~\ref{fig:intra-spanish-qwen} reports ablation outcomes across seven Spanish-speaking national variants. Self-deactivation accuracy drops range from negligible to moderate, with the largest drop for Mexico ($-6.8$), followed by Chile ($-5.1$) and Uruguay ($-2.3$). Flip rates indicate stronger national specificity than accuracy alone: Chile and Uruguay show high self flip rates ($12.0$ and $11.4$), while cross-country flip rates often remain substantial (roughly $7.5$--$12$). This pattern suggests a combination of broadly shared features (e.g., language-linked semantics) and country-distinct cues (e.g., place-specific iconography and landscapes) that can be differentially disrupted by neuron sets identified from different national subsets.

\onecolumn

\section{Complete Unablated Results}
\label{appendix:full_unmasked}
\begin{table}[H]
\centering
\begin{tabular}{lll|ll|ll} 
\toprule
\multicolumn{1}{r}{Model} & \multicolumn{2}{c|}{LLaVA-v1.6-Mistral-7B} & \multicolumn{2}{c|}{Pangea-7B} & \multicolumn{2}{c}{Qwen2.5-VL-7B}  \\
Culture                   & Iden.  & Eval.                             & Iden.  & Eval.                           & Iden.  & Eval.                     \\ 
\midrule
Brazil--Portuguese         & 0.5352 & 0.5493                            & 0.5704 & 0.6338                          & 0.6972 & 0.6901                    \\
Bulgaria--Bulgarian        & 0.4324 & 0.4301                            & 0.5189 & 0.4624                          & 0.6000 & 0.5000                    \\
China--Chinese             & 0.5355 & 0.4679                            & 0.5871 & 0.5641                          & 0.7161 & 0.7308                    \\
Egypt--Egyptian Arabic    & 0.4851 & 0.4608                            & 0.5446 & 0.5294                          & 0.6634 & 0.5686                    \\
Ethiopia--Amharic          & 0.5043 & 0.4957                            & 0.4701 & 0.4530                          & 0.5470 & 0.4274                    \\
Ethiopia--Oromo            & 0.4766 & 0.4486                            & 0.3832 & 0.3551                          & 0.4673 & 0.5794                    \\
France--Breton             & 0.3762 & 0.3202                            & 0.3366 & 0.3202                          & 0.4703 & 0.4039                    \\
India--all                 & 0.5068 & 0.4773                            & 0.6021 & 0.5468                          & 0.6808 & 0.6239                    \\
Indonesia--all             & 0.4594 & 0.3563                            & 0.4801 & 0.4819                          & 0.5250 & 0.5301                    \\
Ireland--Irish             & 0.6074 & 0.5890                            & 0.5644 & 0.6319                          & 0.6196 & 0.6196                    \\
Japan--Japanese            & 0.3069 & 0.3627                            & 0.2178 & 0.2255                          & 0.3861 & 0.4216                    \\
Kenya--Swahili             & 0.4926 & 0.4599                            & 0.4412 & 0.3577                          & 0.5882 & 0.4818                    \\
Malaysia--Malay            & 0.4268 & 0.4304                            & 0.5605 & 0.4873                          & 0.5860 & 0.5380                    \\
Mongolia--Mongolian        & 0.4295 & 0.4295                            & 0.3846 & 0.4167                          & 0.4551 & 0.4679                    \\
Nigeria--Igbo              & 0.5600 & 0.4200                            & 0.4700 & 0.4000                          & 0.5200 & 0.4800                    \\
Norway--Norwegian          & 0.5570 & 0.5133                            & 0.5034 & 0.5133                          & 0.5839 & 0.5800                    \\
Pakistan--Urdu             & 0.5741 & 0.5648                            & 0.3796 & 0.6296                          & 0.7315 & 0.7037                    \\
Philippines--Filipino      & 0.5050 & 0.4804                            & 0.5644 & 0.4804                          & 0.5743 & 0.5882                    \\
Romania--Romanian          & 0.4768 & 0.4570                            & 0.6159 & 0.5563                          & 0.6556 & 0.6556                    \\
Russia--Russian            & 0.3900 & 0.5500                            & 0.5000 & 0.5500                          & 0.5600 & 0.6400                    \\
Rwanda--Kinyarwanda        & 0.4103 & 0.4492                            & 0.3333 & 0.4322                          & 0.4444 & 0.5254                    \\
Singapore--Chinese         & 0.5849 & 0.5094                            & 0.5377 & 0.6038                          & 0.7170 & 0.7358                    \\
South Korea--Korean        & 0.5793 & 0.5034                            & 0.5310 & 0.5517                          & 0.6207 & 0.5655                    \\
all--Spanish               & 0.4557 & 0.5034                            & 0.4830 & 0.4995                          & 0.5871 & 0.5703                    \\
Sri Lanka--Sinhala        & 0.3839 & 0.5310                            & 0.3393 & 0.6195                          & 0.6786 & 0.6726                    \\ 
\midrule
Avg.                      & 0.5345 & \multicolumn{1}{l}{0.5102}        & 0.5938 & \multicolumn{1}{l}{0.5831}      & 0.6522 & 0.6395          \\
\bottomrule
\end{tabular}
\caption{\textbf{Culture-specific performance}. We report accuracy on the set of questions-answer pairs used for neuron identification and evaluation respectively. } %
\label{table:baseline_full}
\end{table}

\begin{table}[h]
\resizebox{\columnwidth}{!}{%
\begin{tabular}{@{}llll@{}}
\toprule
                                                             & Qwen2.5-VL-7B            & \multicolumn{1}{c}{Pangea-7B} & \begin{tabular}[c]{@{}l@{}}LLaVa-v1.6\\ -Mistral-7B\end{tabular} \\ \midrule
Mean of neuron-wise standard deviations across cultures      & 0.015284                 & 0.017388                      & 0.020062                                                         \\
Std. dev. of neuron-wise standard deviations across cultures & 0.012924                 & 0.018586                      & 0.016303                                                         \\
Max neuron-wise standard deviation                           & 0.170701                 & 0.196799                      & 0.160319                                                         \\
Number of neurons where std \textgreater mean activation     & \textbf{65101 (12.27\%)} & \textbf{50772 (9.57\%)}      & \textbf{148 (0.03\%)}                                            \\ \bottomrule
\end{tabular}%
}
\caption{\textbf{Neuron-wise standard deviations across cultures} for the three evaluated models. The preliminarily experiment on Qwen2.5-VL-7B and Pangea-7B yield high variances across cultures.}
\label{table:mad_stat}
\end{table}

\end{document}